\documentclass{article}

\PassOptionsToPackage{numbers, compress}{natbib}

\usepackage[preprint, nonatbib]{neurips_2026}

\usepackage[utf8]{inputenc} %
\usepackage[T1]{fontenc}    %
\usepackage{hyperref}       %
\usepackage{url}            %
\usepackage{booktabs}       %
\usepackage{amsfonts}       %
\usepackage{nicefrac}       %
\usepackage{microtype}      %
\usepackage{xcolor}         %

\usepackage{amsmath}        %
\usepackage{caption}        %
\usepackage{float}          %
\usepackage{graphicx}       %
\usepackage{placeins}       %

\usepackage[
  backend=biber,
  style=numeric,
  sorting=nyt,
  date=year,
  giveninits=true,
  maxbibnames=6,
  doi=false,
  url=false,
  isbn=false,
  eprint=true
]{biblatex}

\AtEveryBibitem{%
  \clearfield{pubstate}%
  \clearfield{eventtitle}%
  \clearfield{eventtitleaddon}%
  \iffieldundef{booktitle}
    {}
    {\clearfield{eprint}}%
  \iffieldundef{journaltitle}
    {}
    {\clearfield{eprint}}%
  \clearfield{eprintclass}%
}

\DeclareFieldFormat[
  article,
  inproceedings,
  misc
]{title}{#1}

\title{Shape-based Figure-Ground Organization in Vision Transformers}

\author{Matthias Tangemann$^{1,2}$ \quad
Benjamin Lo$^{3,4}$ \quad
Zygmunt Pizlo$^{5}$ \quad
Kaleem Siddiqi$^{3,4}$ \\
\textbf{Dirk B. Walther}$^{1}$ \quad
\textbf{Sven Dickinson}$^{1,2}$ \quad \\
\\
$^1$University of Toronto \quad $^2$Vector Institute \quad 
$^3$McGill University \quad 
$^4$MILA \quad
$^5$UC Irvine \\
\\
\texttt{mtangemann@cs.toronto.edu}
}

\begin{document}

\maketitle

\begin{abstract}
Rich perceptual structure emerges in modern vision encoders.
Several studies have shown that the internal representation in many vision transformers separates foreground objects from the background, however, little is known about the visual cues that have been learned by these models.
In this work, we study whether figure-ground organization in a broad range of 25 vision transformers relies on generalizable shape cues, rather than on superficial texture statistics.
To do so we train linear probes to classify patches as figure or ground, and evaluate them zero-shot on synthetic stimuli where individual cues are systematically manipulated.
For a broad range of vision encoders, we find that probes are robust to the random replacement of textures in the input image, suggesting that the emerging figure-ground organization relies on shape cues.
Probes also generalize in a zero-shot fashion to synthetic images where individual shape cues, recognizability and spatial priors are controlled for. 
Our experiments reveal a strong influence of surroundedness and convexity cues, while results for parallelism and symmetry are more nuanced.
More detailed analyses reveal stronger evidence for shape-based cues in larger models, that models trained for object-recognition tend to be less robust to texture changes and that figure-ground information emerges most prominently in middle layers.
Taken together, our study demonstrates that emerging figure-ground organization in recent vision encoders relies strongly on generalizable shape cues, advancing our understanding of the internal processing in vision transformers, and opening an exciting avenue for computational research on human visual perception.

Code and data is available at \href{https://github.com/mtangemann/mlvbench}{https://github.com/mtangemann/mlvbench}.
\end{abstract}

\section{Introduction}
\label{sec:introduction}

Figure-ground organization is an important process in mid-level vision that identifies a foreground object (the \textit{figure}) from its surroundings (the \textit{ground}).
The resulting structured representation forms the basis for higher level tasks such as recognition or spatial reasoning, and enables generalization to novel situations.
For example, adequately responding to an unknown obstacle when driving a car requires correctly separating the object from its surroundings, while recognizing the object is not necessary.
Better understanding figure-ground organization in mid-level representations of vision models therefore greatly contributes to building more robust visual intelligence.

Perceptual structure, including figure-ground organization, has been shown to emerge from end-to-end training in recent vision transformers \cite{caron2021dino,simeoni2026dinov3, bolya2025perception}.
Which image cues these models use to separate figure and ground, however, remains largely unknown.
Research on human visual perception has highlighted the strong influence of shape cues, including many of the classical Gestalt principles \cite{wagemans2012century1}.
Several of these cues have been shown to be rooted in the statistics of natural scenes \cite{geisler2001edge,elder2002ecological}, so it is plausible that vision encoders might acquire similar cues during training.
On the other hand, DNNs have been classically shown to rely primarily on texture cues \cite{geirhos2019imagenettrained,baker2018deep,brendel2019bagnet}.
This raises the question of whether the emerging figure-ground organization is mainly driven by texture cues as well, or whether recent methods have succeeded in learning more Gestalt-like cues.

\begin{figure}[tb]
    \centering
   \includegraphics[width=0.56\linewidth]{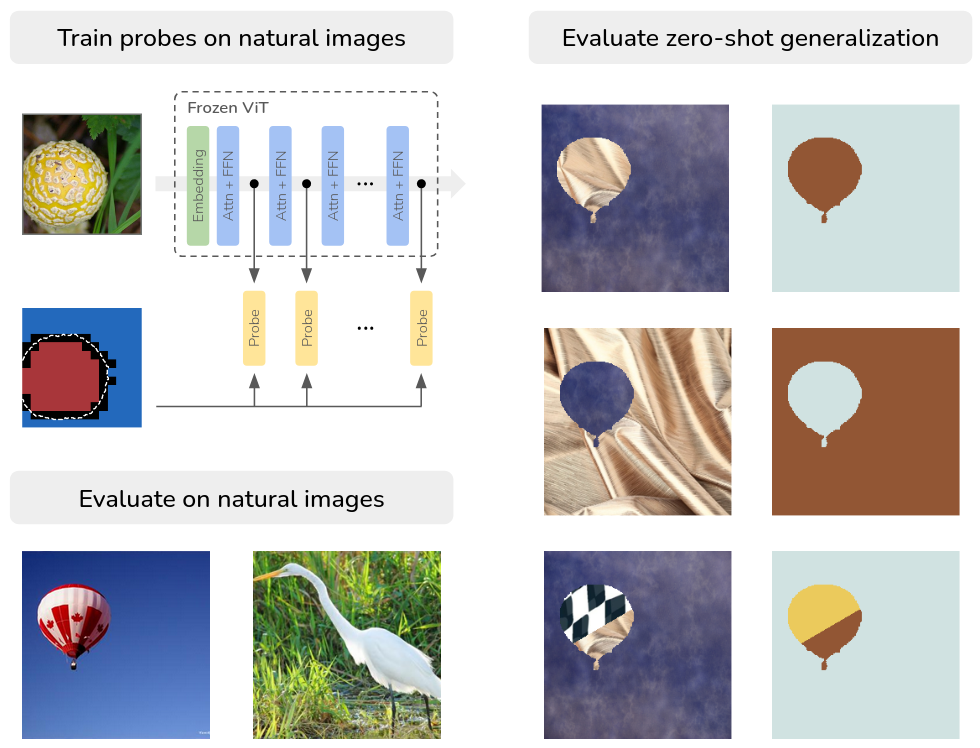}
    \hfill
    \includegraphics[width=0.4\linewidth]{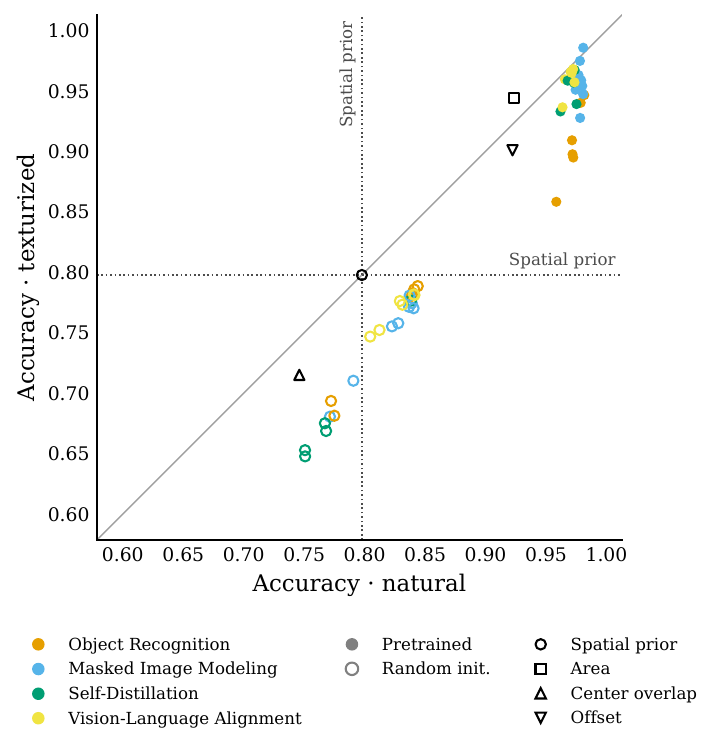}
    \caption{\textit{(left)} We extract patch representations from frozen, pre-trained ViTs and fit linear probes to predict figure-ground assignment on natural images, excluding patches that span both foreground and background (shown in black). We evaluate probes on both natural images and cue-controlled stimuli where textures are randomly replaced. \textit{(right)} Probes for most models are largely unaffected by disrupting texture information, indicating the use of shape cues. Baselines rule out architectural biases and trivial strategies based on region area and location.}
    \label{fig:approach}
\end{figure}

In this work, we systematically study the role of shape cues for emergent figure-ground organization by evaluating the zero-shot generalization of linear probes (Figure~\ref{fig:approach}).
We extract spatial features from pretrained vision transformers and fit linear probes to classify patches as figure or ground on natural images.
Testing the zero-shot generalization of natural image probes to cue-controlled experimental stimuli then allows us to draw conclusions about the figure-ground cues used by the network. 
If the model relies primarily on texture cues, probes trained on natural images will not generalize to stimuli in which all textures are randomly replaced.
Conversely, if the network relies on shape cues, we would expect the probe performance to be unaffected.
Following the same logic, we further test the use of individual shape cues discussed in human vision science: surroundedness, convexity, symmetry and parallelism~\cite{wagemans2012century1}.

Our results show that shape-based cues for figure-ground organization emerge in a broad range of 25 vision encoders.
All tested models encode figure-ground information for natural scenes and show some level of robustness to texture disruptions, clearly outperforming random features and trivial baselines based on region area and location.
A subset of models achieves the same performance on randomly textured images as on the original test set, and shows strong biases towards surroundedness and convexity.
More detailed analyses further reveal that larger models tend to show stronger use of shape cues while we do not observe a clear influence of the pretraining task.
These results demonstrate that the emerging figure-ground organization in many vision transformers relies on generalizable shape cues that have been learned during pretraining.

In summary, our paper makes the following contributions:

\begin{itemize}
\item We show that figure-ground information emerges across a broad range of 25 Vision Transformers pretrained on different tasks.
\item We demonstrate that figure-ground organization strongly relies on shape cues for most models, and in particular surroundedness and convexity cues.
\item We find that figure-ground information typically peaks in middle layers and that model size and the pretraining task influence the emergence of shape cues.
\end{itemize}

Our findings represent an important contribution towards understanding the emerging representations in vision transformers.
To our knowledge, we are the first to demonstrate the role of shape cues for emergent figure-ground organization.
This qualitative similarity to classical Gestalt cues provides an exciting direction for computational research on perceptual organization in human vision.

\section{Related Work}
\label{sec:related-work}

\paragraph{Figure-ground organization.}
Figure-ground organization is a central capability of the human visual system that contributes to structuring the sensory input into coherent surfaces and objects \cite{wagemans2012century1,palmer1999vision,wolfe2021sensation}.
Research in psychophysics has identified a range of cues that are used by the human visual system for figure-ground assignment (\cite{rubin1921visuell,bahnsen1928untersuchung,kanisza1976convexity,stevens1988concave,peterson1994objecta,peterson2008inhibitory} see \cite{wagemans2012century1} for a review).
Among these, surroundedness, convexity, and symmetry play central roles in figure-ground assignment and shape interpretation. Regions that are spatially surrounded are more likely to be perceived as figure rather than background, while convex regions tend to dominate over concave counterparts in determining object boundaries. Symmetry further biases perception by promoting the grouping of elements into unified forms, reflecting the visual system’s sensitivity to regularity and structural simplicity.

\paragraph{Emerging perceptual organization in vision transformers.}
Caron et al. \cite{caron2021dino} observed that object-centric structure emerges in Vision Transformers trained using the self-supervised DINO method.
Following models like iBOT \cite{zhou2021image} and improved versions of the DINO model \cite{oquab2024dinov2,darcet2024vision,simeoni2026dinov3} further confirmed this observation.
This inspired a range of segmentation methods based on clustering patch features from pretrained ViTs (\cite{simeoni2021lost,hamilton2022stego,melas-kyriazi2022spectral,wang2022tokencut,amir2023effectiveness,simeoni2023unsupervised,rambhatla2023most,zhang2024heap,wu2025ensemble,rawlekar2026finding}, see \cite{simeoni2025unsupervised} for a review).
Using the emergent segmentation as pseudo-ground truth has further enabled unsupervised segmentation methods to approach the performance of their supervised counterparts \cite{wang2023cutler,wang2024unsam,yu2026unsamv2}.
While DINO models remain the most popular models for studying emergent perceptual organization, other works have demonstrated the emergence of object segmentation in a broader range of vision transformers \cite{couairon2024diffcut,li2025objectbinding}.

\paragraph{Shape-based processing in vision encoders.}
A long line of work has studied whether DNNs learn shape-based visual processing similar to the human visual system (see \cite{wichmann2023adequate} for a recent review).
One line of work has revealed that object recognition CNNs are strongly driven by texture cues, while humans dominantly rely on global shape \cite{baker2018deep,geirhos2018generalisation}.
Moreover, early DNNs could be approximated quite well by networks that only consider local texture information \cite{brendel2019bagnet}.
This trend however seems to be changing, as more recent studies increasingly find evidence for more human aligned, shape-based processing in vision models \cite{geirhos2021partial,dehghani2023scaling,burgert2025imagenettrained}.
Recently, \cite{adeli2026humanlike} have further shown that the emerging segmentation in self-supervised vision encoders aligns with human object perception.

Several studies have investigated whether DNNs follow classical Gestalt principles of perceptual grouping, including both failure and success cases depending on the model and experimental paradigm \cite{baker2018illusionary,amanatiadis2018understanding,ehrensperger2019evaluating,kim2021neural,biscione2023mixed,tang2023degraded,zhang2024investigating,li2025gestalt,zhang2025finding,lonnqvist2025contour,li2025gestalt,sha2026gestalt,tartaglini2026walk,li2026learning}.
Far fewer works have studied whether figure-ground organization in vision encoders follows similar cues as the human visual system, suggesting a role of convexity in masked image models \cite{li2025gestalt,arnold2026locating}.

\section{Methods}
\label{sec:methods}
We examine whether pretrained ViTs use shape cues for figure-ground organization by studying the generalization of linear probes to cue-controlled stimuli.
All probes are trained on natural images to extract figure-ground information from frozen ViT features.
Evaluating the zero-shot generalization of these probes to cue-controlled stimuli then allows us to draw conclusions about the cues that the model relies on:
Probes for models that primarily rely on texture rather than shape cues will not generalize from natural images to stimuli where the textures are randomly replaced.
Below, we describe the protocol for probe fitting and evaluation (Section~\ref{sec:probes}) and the vision encoders considered in our study (Section~\ref{sec:models}).
The cue-controlled test conditions are described with the experiments in Section~\ref{sec:experiments}.

\subsection{Probe Fitting and Evaluation}
\label{sec:probes}

We fit linear figure-ground probes on images and ground truth segmentation masks from the MSRA-10K dataset \cite{cheng2015global}, which provides photographs featuring a single, dominant foreground object. The dataset is randomly split into 80\% training, 10\% validation, and 10\% test images. Each image is resized so that the shorter side matches the model's native input resolution and is then center-cropped.

The probed ViT is applied to the input images $x \in \mathbb{R}^{H\times W\times 3}$, and intermediate features $f^{(l)} \in \mathbb{R}^{h\times w\times C}$ are extracted for each layer $l$, where $(h,w) = (H/p, W/p)$ and $p$ is the patch size of the model. We downscale the ground truth segmentation masks $s \in \{0,1\}^{H\times W}$ to the internal grid, yielding $\tilde{s} \in \{0,1\}^{h\times w}$. Patches that overlap both foreground and background regions receive ambiguous labels and are therefore masked in the training loss and evaluation metrics.

\paragraph{Probes.}
For each model layer, we fit a logistic regression model that predicts the probability of a patch belonging to the foreground:

$$
    \text{probe}(f_{i,j}) = \sigma(w^Tf_{i,j} + b)
$$

Probes are trained with binary cross-entropy (BCE) loss using the Adam optimizer \cite{kingma2015adam} with a batch size of 256 for 2000 steps, a learning rate of $0.001$ and no weight decay. Inspection of validation loss curves confirmed that probes converge within this limit. We evaluate probes every 200 steps, restore the best step and select the best layer based on the validation loss. All results are reported on held-out test sets. In total, we fit 3108 probes on combined 1.6B samples within 120 GPU hours (NVIDIA L40S).

\paragraph{Evaluation.}
We evaluate all probes on the held-out i.i.d. test set, and apply them zero-shot to several cue-controlled conditions. We use accuracy as the primary metric, and additionally report BCE, information gain over the spatial prior, and intersection-over-union (IoU) in the appendix.

\paragraph{Baselines.}
We consider several baselines to rule out trivial explanations for the observed results:

\begin{itemize}
    \item \textit{Spatial prior}: We compute the prior foreground probability for each location $(i,j)$ on the training set and evaluate this constant prediction on all test sets. This corresponds to a probe that purely relies on the positional information in ViT features.
    \item \textit{Random features}: We fit probes for randomly initialized variants for all models using the same protocol as described above in order to rule out any architectural biases.
    \item \textit{Region properties}: We use ground truth texture segmentation information and select the foreground region based on heuristic properties including the region area, overlap with the image center, and the offset of the region centroid from the image center. This simulates models that apply texture segmentation in combination with a trivial heuristic.
\end{itemize}

\subsection{Models}
\label{sec:models}
We study a diverse set of 25 Vision Transformer models. For all models, we use the implementations and checkpoints as provided by the \texttt{timm} library~\cite{wightman2019timm} and test the ViT-B and ViT-L variants. We consider ViTs trained for supervised and unsupervised training objectives:

\begin{itemize}
\item \textbf{Object recognition} models are trained with supervision on ImageNet \cite{deng2009imagenet}. We include an improved version of the original ViT (AugReg, \cite{steiner2022how}), DeIT III \cite{touvron2022deit3} and FlexiViT \cite{beyer2023flexivit}.

\item \textbf{Vision-Language Alignment} models have been pioneered by CLIP \cite{radford2021clip}. We additionally consider the more recent SigLIP 2 \cite{tschannen2025siglip2} and Perception Encoder Spatial models \cite{bolya2025perception}.

\item \textbf{Masked Image Modeling} is a self-supervised pretraining task where the model is trained to reconstruct missing patches in the input image. We include the original Masked Auteoncoder \cite{he2022masked} and BEiT \cite{bao2021beit}. Further, we consider EVA-02 \cite{fang2024eva02}, which is trained to reconstruct CLIP features for masked patches, and BEiT-3 \cite{wang2023image}, which is jointly trained on masked images and text and finetuned on ImageNet. 

\item \textbf{Self-Distillation} is a self-supervised pretext task used by the DINO models. We include the original DINO model, DINOv2 with registers, and DINOv3 \cite{caron2021dino,oquab2024dinov2,darcet2024vision,simeoni2026dinov3}.
\end{itemize}

Links to the precise model checkpoints are provided in Table~\ref{tab:model-info} in the appendix.

\section{Experiments}
\label{sec:experiments}

\begin{figure}[tb]
    \centering
    \includegraphics[width=\linewidth]{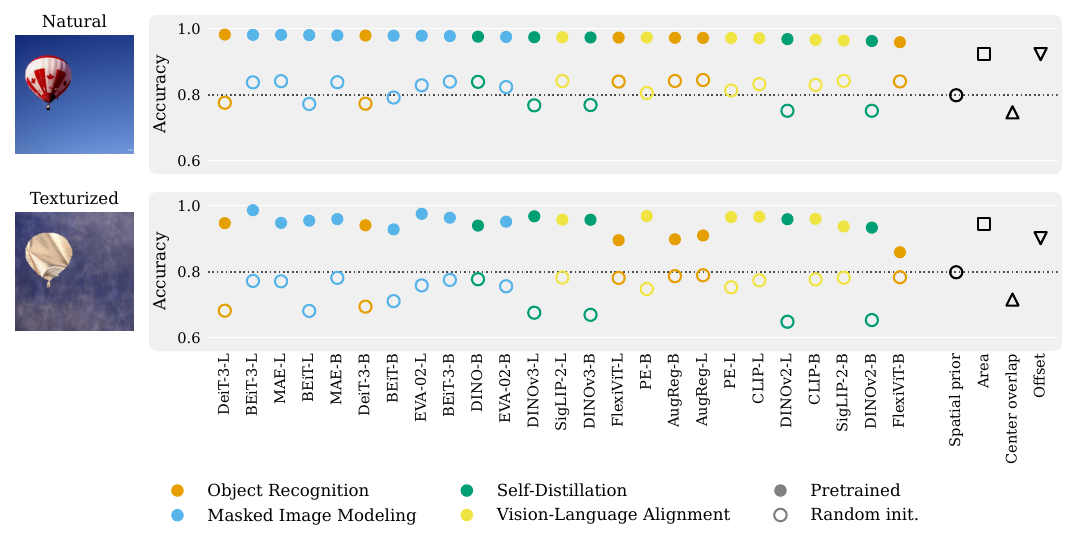}
    \caption{Probe performance for the i.i.d. natural test set, and applied zero-shot to the same images but with the textures randomly replaced. Many models reach high accuracies for both the natural and texturized images, demonstrating the use of shape cues in these models.}
    \label{fig:overall-performance}
\end{figure}

All experiments are based on linear probes that are trained to classify image patches as figure or ground in natural scenes.
The performance of the models on the i.i.d. test set in Figure~\ref{fig:overall-performance} shows that patch tokens from all 25 tested models provide strong figure-ground information.
The results further reveal a substantial spatial bias, which is likely caused by the tendency of photographs to contain the foreground object near the image center.
While the probes for pretrained models outperform this spatial prior by a large margin, probes for random features perform much worse.
Figure-ground organization is therefore not an architectural bias but requires learning.

\begin{figure}[tb]
    \centering
    \includegraphics[width=\linewidth]{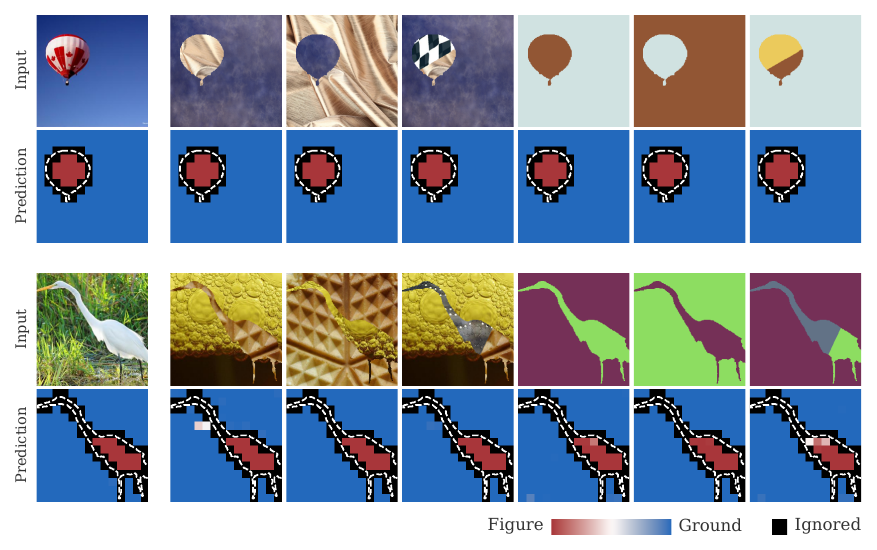}
    \caption{Figure-ground predictions based on layer 12 from BeiT-3-L \cite{wang2023image}. The probe generalizes from natural images to randomized textures with hardly any loss in performance, providing support for shape-based figure-ground cues used by the model.}
    \label{fig:predictions-single-model}
\end{figure}

So far, our results corroborate previous studies showing the emergence of perceptual structure in Vision Transformers and further extend the range of tested models.
In order to gain insight about the visual cues that underlie this phenomenon, we assess whether figure-ground probes trained on natural images generalize in a zero-shot fashion to stimuli in which shape and texture cues are systematically controlled.
In Section~\ref{sec:experiments-generalization} we test whether models are robust to the disruption of texture cues, while in Section~\ref{sec:experiments-cues} we take a more detailed look at individual shape cues.
Finally, in Section~\ref{sec:experiments-factors} we evaluate the influence of model size and pretraining task, and in Section~\ref{sec:experiments-layers} we test which layers provide the strongest figure-ground information.

\subsection{Generalization to texturized stimuli}
\label{sec:experiments-generalization}

If figure-ground organization in Vision Transformers relies on shape cues, we would expect the probe predictions to be invariant to changes in texture.
Conversely, if figure-ground organization relies purely on texture cues, we would expect the performance to drop to chance level.
To test these possibilities, we generate a variant of the natural test set with the same segmentation mask, but where each segment is filled with a random texture from the DTD dataset \cite{cimpoi2014describing} or a random color.
In total, we create 6 variants for each image (see Figure~\ref{fig:approach} for examples). 
We fill figure and ground with random textures or colors, and as a control include variants where the assignment is reversed.
We observed that the performance of baselines using texture segmentation with a trivial foreground heuristic was quite high for these variants.
We therefore include additional stimulus variations where the figure is split by a straight line and filled with two different textures or colors.
All results reported in the main paper average over the different variants, and more detailed results are included in the appendix.

The results in Figure~\ref{fig:overall-performance} provide strong evidence for shape-based figure-ground organization in many Vision Transformers, as probes from most models generalize well to synthetic stimulus variants where texture information is destroyed.
The direct comparison of the i.i.d. and zero-shot performances in Figure~\ref{fig:approach} further shows that for several models, probes generalize to changes in texture without any loss in performance.
For other models, and in particular object recognition models except DeiT-III \cite{touvron2022deit3}, we observe a substantial drop in performance, yet there is still a clear improvement over the spatial prior baseline.
The results for the individual variants (Figure~\ref{fig:natural-vs-texturized-variants} in the appendix) show that these models struggle on the colorized stimuli, while they are less affected for the textured variants.

\subsection{Isolation of shape cues}
\label{sec:experiments-cues}

\begin{figure}[tb]
    \centering
    \includegraphics[width=\linewidth]{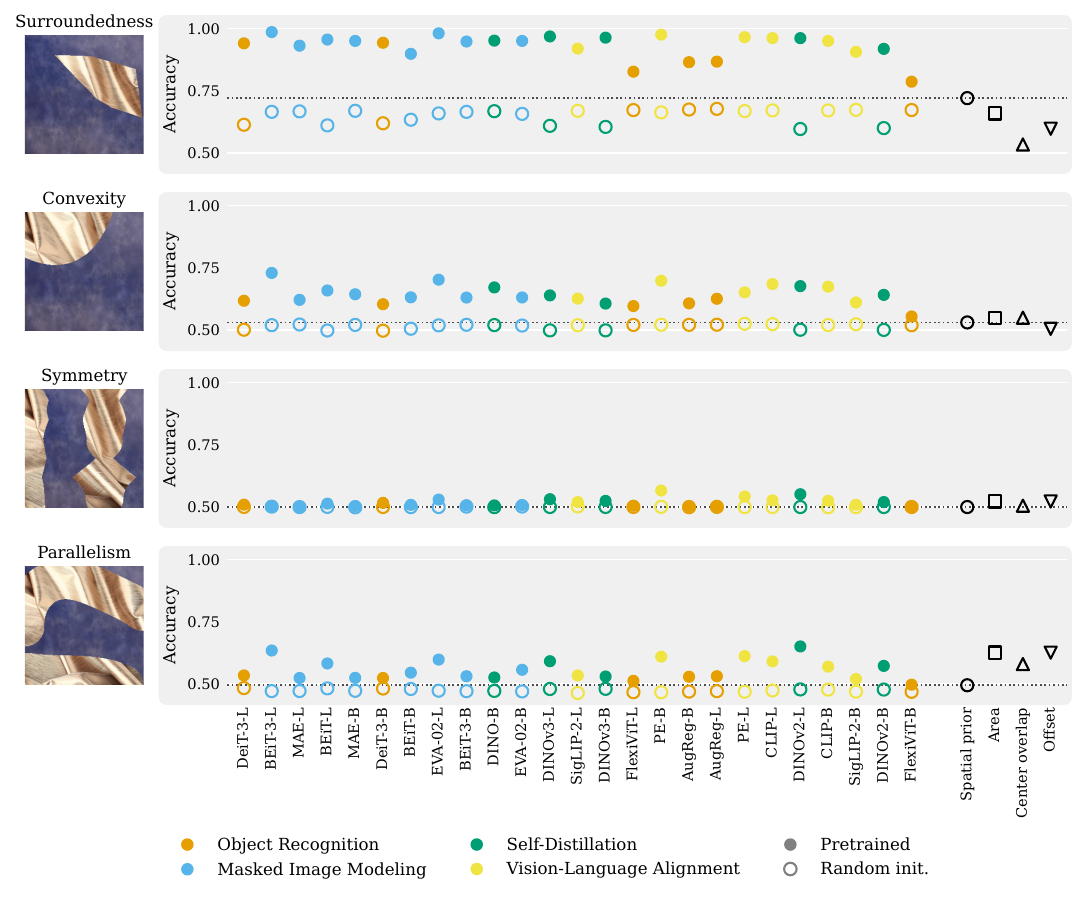}
    \caption{Zero-shot performance of the figure-ground probes on cue-controlled test sets. The results shows strong biases towards surrounded and convex regions, a weaker role of parallelism and no clear evidence for the use of symmetry.}
    \label{fig:performances-cues}
\end{figure}

The previous results demonstrate that figure-ground organization in many ViTs relies on shape cues.
However, we do not know \textit{which} shape cues have been learned by the models.
Moreover, due to the strong spatial bias in natural images, we cannot fully rule out a strong influence of trivial positional heuristics.
In this section, we therefore test the figure-ground probes zero-shot on stimuli designed to isolate the role of individual shape cues.
We focus on four cues that have been discussed in the human vision literature: surroundedness, convexity, parallelism and symmetry (see examples in Figure~\ref{fig:performances-cues}).

\begin{itemize}
    \item \textit{Surroundedness}: Stimuli contain a random shape from the idsprites dataset (\cite{dziadzio2024infinite}). The shapes include both convex and concave regions, and are surrounded by the background on all sides. The relative area covered by the shape is uniformly sampled from $[0.08, 0.25]$ and $[0.5, 0.6]$, respectively, where both intervals are chosen with equal probability. When selecting the first interval, we additionally ensure that the image center is not covered by the shapes.
    \item \textit{Convexity}: Foreground and background are separated by a random parabolic arc, $v = ku^2 +c$ in a randomly oriented coordinate frame, with $k \sim \text{Uniform}(1.0, 3.0)$ and $c$ chosen to ensure a specific region area. We randomly sample the region area in $[0.2, 0.3]$ and $[0.7, 0.8]$, respectively, and both intervals are selected with equal probability.
    \item \textit{Symmetry}: The image is divided into four equally sized vertical columns by three Bézier-curve edges. Two consecutive edges are made mirror-symmetric by copying and inverting their control points, producing two symmetric columns (foreground) flanked by two asym- metric columns (background). This design follows the classic Bahnsen column paradigm \cite{bahnsen1928untersuchung}.
    \item \textit{Parallelism}: Stimuli contain random stripes with a constant thickness of one third of the stimulus width. We generate the stripes by sampling curvature values from a Gaussian process prior with a squared exponential kernel and integrating the sampled function, starting from a random point on the image boundary.
\end{itemize}

The parameters for all stimulus classes, including the size ranges, have been selected in order to minimize the accuracy of the simple center-related baselines.
We use random textures and colors for the regions, including the reverse variants.
For the surroundedness and convexity stimuli, we additionally consider variants where the foreground region is split into two subregions.

The results in Figure~\ref{fig:performances-cues} show clear evidence for surroundedness and convexity cues in most models.
Probes for all ViTs outperform the baselines in the surroundedness condition, with near perfect predictions in many cases.
The results in the convexity condition are weaker, but we observe a clear bias towards the convex region in most models that cannot be explained by any of the baselines.
In the parallelism condition, we see a bias towards the stripes with parallel outlines, but the results require more careful interpretation. Since some of the baselines achieve similar accuracies, we cannot fully exclude the possibility that models rely on trivial heuristics.
However, the layer-wise analysis in Section~\ref{sec:experiments-layers} shows that for certain models, later layers than those selected on the natural test set are substantially more biased towards parallelism, which provides evidence against the baselines in these cases.
Finally, for the symmetry condition, we do not observe a clear bias towards perceiving the symmetric region as foreground.

Overall, these results contribute further evidence for generic, shape-based figure-ground organization in several Vision Transformers.
The generalization to the surroundedness and convexity conditions demonstrates that figure-ground organization does not depend on texture cues, recognizable shapes, or the strong center bias of natural images.

\subsection{Factors influencing the emergence of shape cues}
\label{sec:experiments-factors}

\begin{figure}[tbh]
    \centering
    \includegraphics[width=\linewidth]{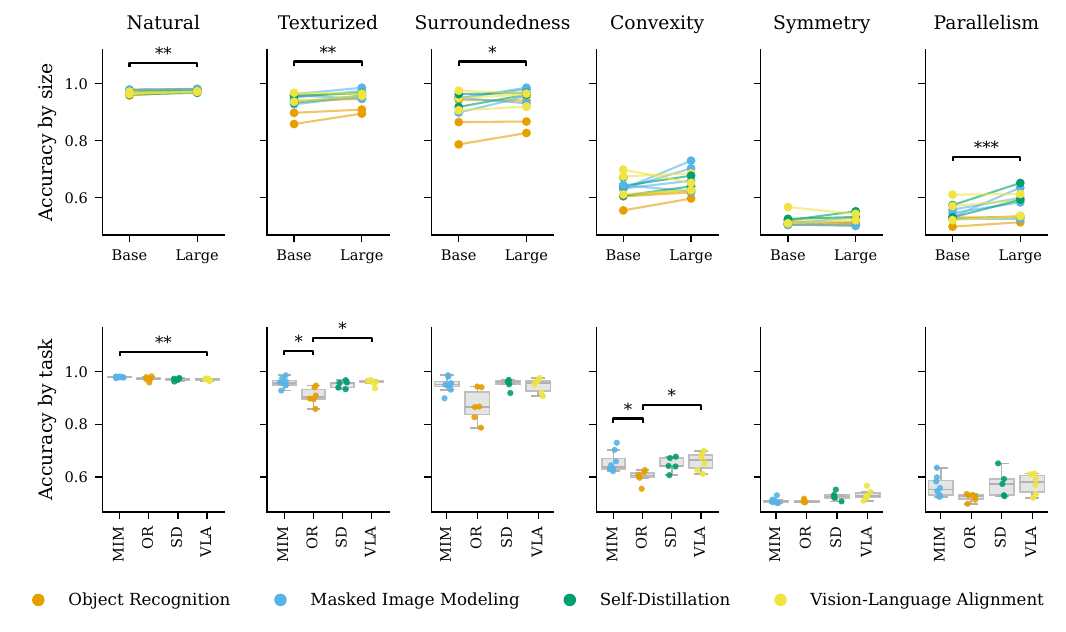}
    \caption{Probe performance by model size (top row) and pretraining task (bottom row: OR = Object Recognition, VLA = Vision-Language Alignment, MIM = Masked Image Modeling, SD = Self-Distillation). Brackets indicate statistical significance: (*/**/***: $p<.05/.01/.001$, non-significant pairs are omitted for clarity).}
    \label{fig:performance-by-size-and-task}
\end{figure}

In Figure~\ref{fig:performance-by-size-and-task}, we analyze which model attributes influence the emergence of shape-based figure-ground organization in ViTs.
We find effects of both model size and pretraining task.
In all cases, except for convexity and symmetry, probes for ViT-L models perform significantly better than probes for the smaller ViT-B models.
Furthermore, we see significant differences between pretraining tasks in many cases, suggesting a stronger overall performance of Masked Image Models, and a weaker overall performance of object recognition models.
However, there are exceptions, as the DeiT-III object recognition model performs substantially better than other object recognition models.

\subsection{Layer-wise analysis}
\label{sec:experiments-layers}

\begin{figure}[tbh]
    \centering
    \includegraphics[width=\linewidth]{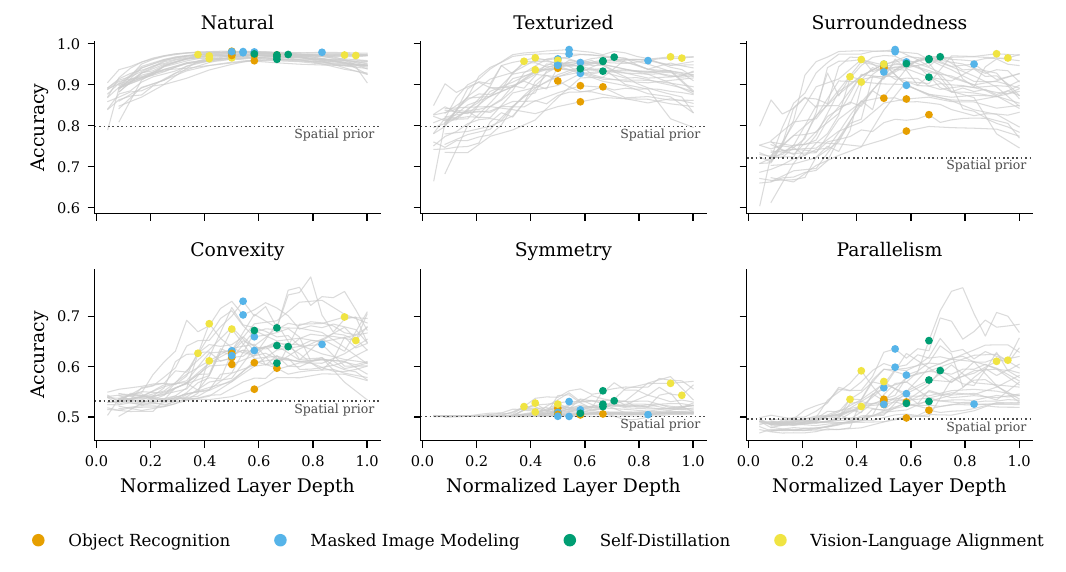}
    \caption{Per-layer performance for each test condition. The best layer selected on the validation set is marked with a colored dot. For most models, figure-ground information is best decoded from intermediate layers. More detailed information is provided in Figure~\ref{fig:performance-by-layer-details} in the appendix.}
    \label{fig:best-layer}
\end{figure}

So far, we have analysed the performance of the best layer for each model, selected based on the validation loss.
In Figure~\ref{fig:best-layer}, we take a more detailed look at the performance of the individual layers.
On natural images, we see a rise in figure-ground information during the first few layers, with a peak around the middle.
For some models, we observe a drop in performance in later layers.

While early layers enable non-trivial predictions for natural images, the probes for early layers perform around chance-level for the generalization conditions.
This suggests that the information extracted by the probes corresponds to features that are superficially correlated but do not provide robust figure-ground information.
In some other cases, and in particular for the convexity and parallelism conditions, it seems that later layers are more strongly biased towards these cues than the layers selected on the validation set.
Taking into account the relatively flat curves in the natural conditions, this is compatible with the hypothesis that figure-ground information plateaus early for the easier natural images but converges later for the more difficult synthetic conditions.

\subsection{Comparing generalization and decodability}
\label{sec:experiments-generalization-vs-decodability}

\begin{figure}[tbh]
    \centering
    \includegraphics[width=\linewidth]{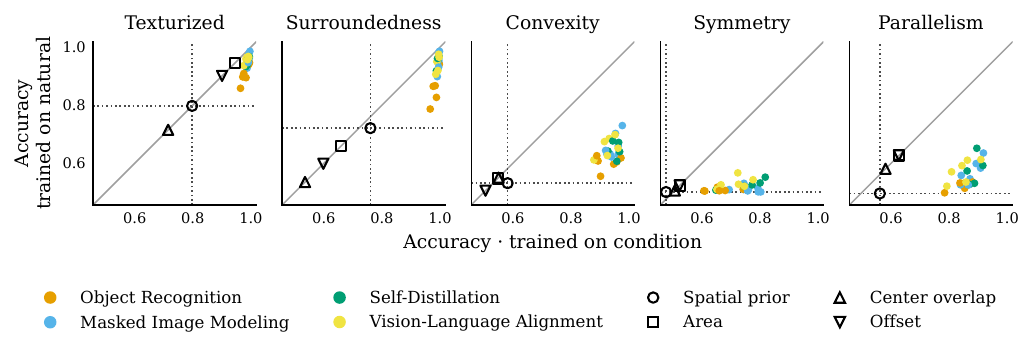}
    \caption{Comparison of probes trained on natural images with probes trained directly on the cue controlled conditions. For the strongest models, direct training on the texturized or surroundedness condition hardly improves over the zero-shot results. The performances improve substantially for convexity, symmetry and parallelism which suggests a weaker link to figure-ground organization.}
    \label{fig:generalization-vs-decodability}
\end{figure}

We train probes directly on training sets matching the cue-controlled stimuli in order to estimate how well the cues can be decoded from the network representations \textit{in principle}. The results in Figure~\ref{fig:generalization-vs-decodability} indicate that several cues, and in particular convexity, symmetry and parallelism, can be decoded much better from internal features than the zero-shot performance of the figure-ground probes. This is the case even for the best performing vision encoders where training directly on texturized images or the surroundedness condition does not improve results. Overall, this suggests that convexity, symmetry and parallelism are less strongly linked to figure-ground organization in these models than surroundedness. 

\section{Limitations}
We rely on patch-wise, linear probes to decode figure-ground information from intermediate ViT features.
This ensures that any positive result reflects readily accessible information that can be used \textit{in principle} by the key, query and value projections in subsequent layers.
In cases where we did not find strong shape-based figure-ground information, it is nevertheless possible that a more powerful probe could successfully decode figure-ground assignment.
Furthermore, while this paradigm allows us to study the cues that influence figure-ground information, our paradigm does not support any conclusions about how figure-ground information influences any higher-level capabilities of the respective network.

Finally, we do not directly compare DNNs with human behavior.
While our study provides qualitative evidence for cues that are believed to underlie figure-ground information in humans as well, we cannot make strong claims about the similarity to human vision.
Since our work focuses on establishing the use of shape cues for perceptual organization in DNNs, this question is out of scope for the present paper.
However, given the qualitative similarity of some of the cues to well known Gestalt principles, this provides a highly promising direction for future work.

\section{Discussion}
We performed a systematic study evaluating whether emergent figure-ground organization in recent Vision Transformers relies on shape cues.
Our experimental approach is based on testing zero-shot generalization of linear probes: only if the emergent structure in ViTs relies on shape cues would we expect the zeros-shot generalization of probes to stimuli where textures are randomly replaced.
The results demonstrate strong generalization for the majority of models, and therefore strongly indicate shape-based processing in ViTs.

This finding challenges the common characterization of deep neural networks as primarily texture-driven.
Classical work has shown strong texture biases of early CNNs \cite{geirhos2019imagenettrained,baker2018deep,brendel2019bagnet}, marking a striking difference to human visual perception.
Our work strongly corroborates more recent studies that suggest a stronger shape bias in more recent models \cite{geirhos2021partial,dehghani2023scaling} since probe performance was not impacted by randomizing textures in several models.
Furthermore, the successful generalization to stimuli with randomly generated shapes rules out a central role of recognizable outlines and spatial biases, while the random feature control rules out architectural biases.
While we cannot conclude that texture cues are never used by the models, our results show that shape-based processing is sufficient for figure-ground organization.
The emergence of these cues across supervised, self-supervised, and vision-language models suggests that learning shape-based figure-ground cues is a robust phenomenon rather than an artifact of any particular training paradigm.

Finally, our results open an exciting avenue for research on human visual perception.
Our results show that several ViTs learned to use human-like cues, including surroundedness and convexity and, to a lesser degree, parallelism.
While we do not find evidence for the use of symmetry, the effectiveness of this cue has been debated for humans as well \cite{kanisza1976convexity,mojica2014displaywide}.
Qualitatively, these results therefore seem compatible with human vision, but direct comparisons will be necessary to make stronger claims.
If these similarities hold, modern vision encoders would be highly valuable model systems for studying open questions about the mechanisms and learning of Gestalt principles in human vision.

\begin{ack}
This work was supported by the Natural Sciences and Engineering Research Council of Canada (NSERC) and Samsung. The authors thank the Digital Research Alliance of Canada (alliancecan.ca) for providing computing resources.
\end{ack}

\printbibliography

\newpage
\appendix

\section{Detailed results}
\label{sec:appendix-detailed-results}

We report the following evaluation metrics in the extended results:

\begin{itemize}
    \item \textbf{Accuracy (ACC)} is computed by thresholding the predicted foreground probabilities for each patch at $0.5$ and comparing to the binary ground truth.
    \item \textbf{Binary cross entropy (BCE)} between the predicted foreground probabilities and the binary ground truth.
    \item \textbf{Information gain (IG)} is defined as the reduction of the binary cross entropy relative to the spatial prior:
    $$
        \text{IG} = \text{BCE}_{prior} - \text{BCE}_{probe}
    $$
    \item \textbf{Information gain explained (IGE)} is the fraction of the maximum possible information gain achieved by the respective probe.
    $$
    \text{IGE} = \frac{\text{IG}}{\text{BCE}_{prior}} = 1 - \frac{\text{BCE}_{probe}}{\text{BCE}_{prior}}
    $$
    An IGE of 0\% indicates that the probe performs no better than the spatial prior, while an IGE of 100\% indicates perfect prediction.
    \item \textbf{Mean intersection over union (mIoU)} is computed for the predictions thresholded at $0.5$. For each sample, we compute the IoU between the predicted and target foreground and background regions, respectively, and average their scores.
\end{itemize}

The region property baselines predict hard binary masks so that the binary cross entropy is not finite for incorrect predictions. We therefore use accuracy as the main metric throughout the paper.

\clearpage

\subsection{Natural stimuli}

\begin{figure}[H]
    \centering
    \includegraphics[width=0.6\linewidth]{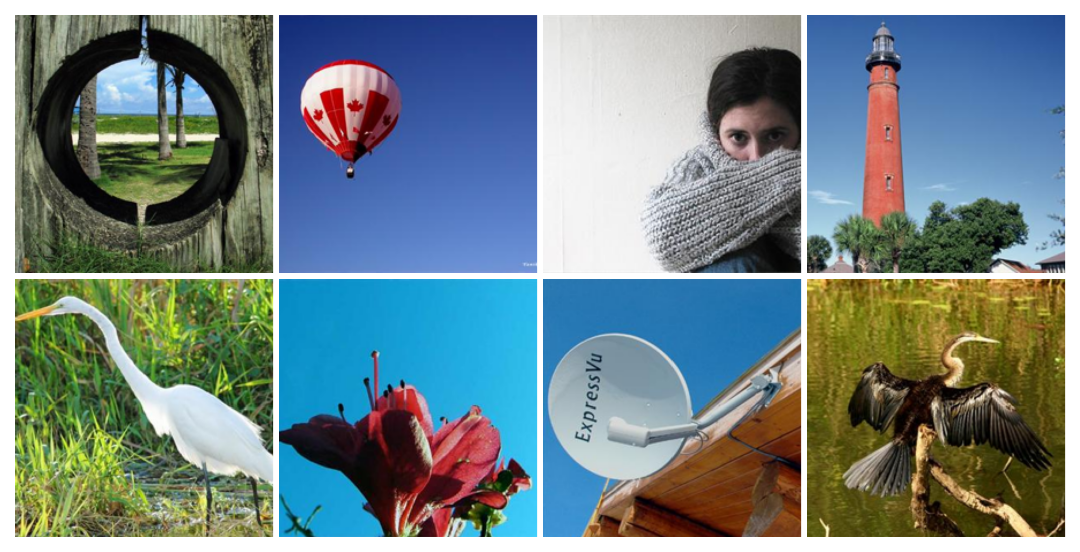}
    \caption{Example images from the natural test set (MSRA-10K, \cite{cheng2015global})}
    \label{fig:examples-natural}
\end{figure}

\begin{table}[H]
  \begin{table}[H]
\centering
\small
\caption{Probe performances on the natural test set.}
\label{tab:performance-natural}
\vspace{7pt}
\begin{tabular}{l|rrrrr}
Model & ACC $\downarrow$ & BCE & IG & IGE & mIoU \\
\hline
DeiT-3-L & 0.982 & 0.051 & 0.360 & 0.876 & 0.951 \\
BEiT-3-L & 0.981 & 0.054 & 0.358 & 0.870 & 0.950 \\
MAE-L & 0.981 & 0.054 & 0.357 & 0.869 & 0.951 \\
BEiT-L & 0.980 & 0.056 & 0.356 & 0.865 & 0.948 \\
MAE-B & 0.979 & 0.058 & 0.354 & 0.860 & 0.947 \\
DeiT-3-B & 0.979 & 0.059 & 0.352 & 0.856 & 0.946 \\
BEiT-B & 0.978 & 0.060 & 0.351 & 0.853 & 0.945 \\
EVA-02-L & 0.978 & 0.061 & 0.359 & 0.855 & 0.945 \\
BEiT-3-B & 0.977 & 0.062 & 0.349 & 0.848 & 0.941 \\
DINO-B & 0.975 & 0.068 & 0.344 & 0.835 & 0.939 \\
EVA-02-B & 0.974 & 0.071 & 0.349 & 0.831 & 0.937 \\
DINOv3-L & 0.974 & 0.072 & 0.348 & 0.829 & 0.935 \\
SigLIP-2-L & 0.974 & 0.075 & 0.345 & 0.821 & 0.934 \\
DINOv3-B & 0.973 & 0.074 & 0.345 & 0.823 & 0.935 \\
FlexiViT-L & 0.973 & 0.073 & 0.343 & 0.824 & 0.931 \\
PE-B & 0.972 & 0.078 & 0.376 & 0.829 & 0.937 \\
AugReg-B & 0.972 & 0.080 & 0.331 & 0.806 & 0.930 \\
AugReg-L & 0.971 & 0.082 & 0.330 & 0.801 & 0.930 \\
PE-L & 0.971 & 0.081 & 0.372 & 0.821 & 0.934 \\
CLIP-L & 0.970 & 0.080 & 0.341 & 0.810 & 0.926 \\
DINOv2-L & 0.968 & 0.088 & 0.370 & 0.809 & 0.926 \\
CLIP-B & 0.966 & 0.095 & 0.317 & 0.770 & 0.915 \\
SigLIP-2-B & 0.964 & 0.104 & 0.307 & 0.746 & 0.908 \\
DINOv2-B & 0.962 & 0.102 & 0.356 & 0.777 & 0.913 \\
FlexiViT-B & 0.958 & 0.112 & 0.304 & 0.731 & 0.898 \\
\hline
Spatial prior & 0.798 & 0.433 & 0.000 & 0.000 & 0.571 \\
Area & 0.924 & -- & -- & -- & -- \\
Center overlap & 0.746 & -- & -- & -- & -- \\
Offset & 0.922 & -- & -- & -- & -- \\
\end{tabular}
\end{table}

\end{table}

\begin{figure}[H]
    \centering
    \includegraphics[width=\linewidth]{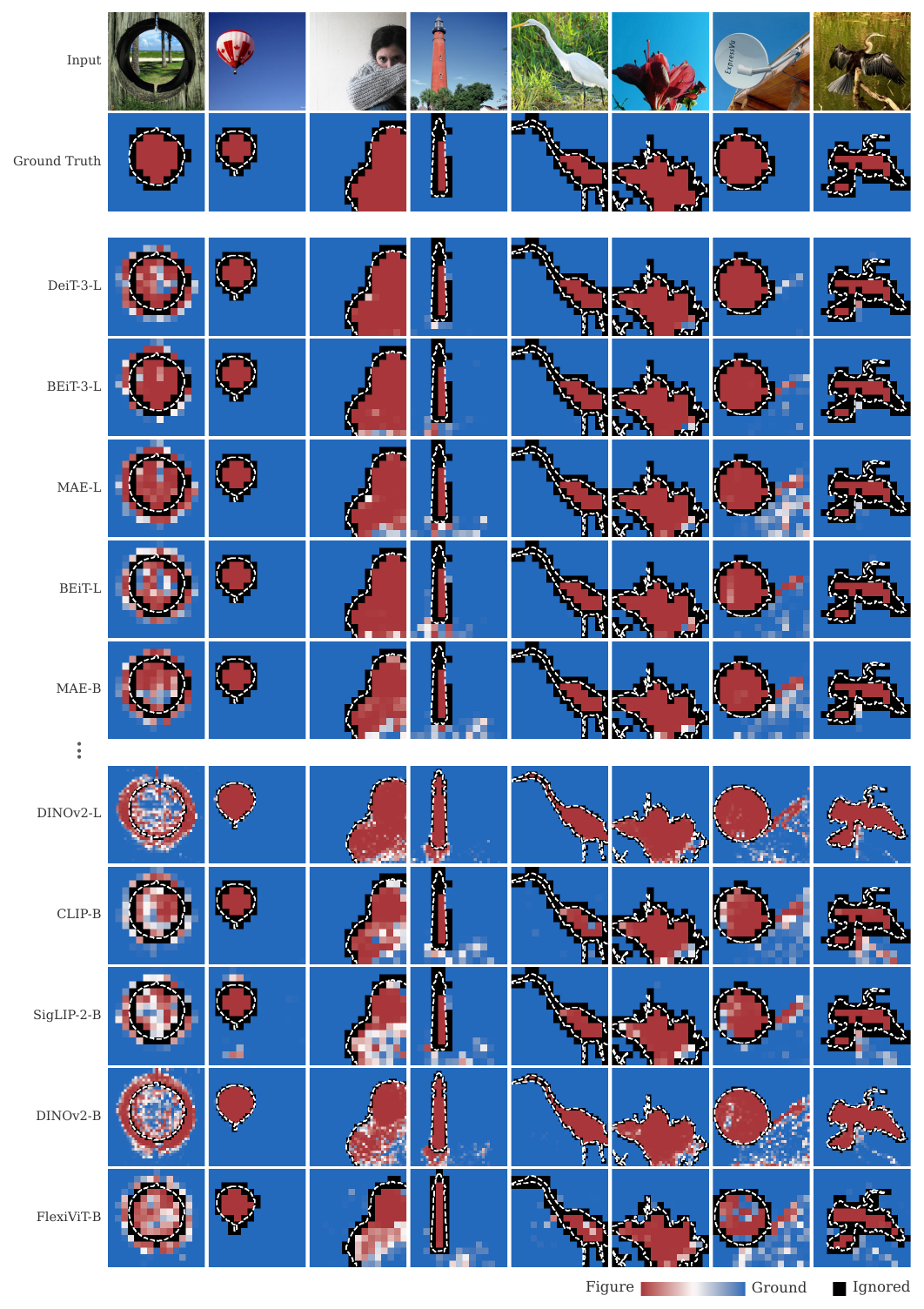}
    \caption{Probe predictions for the best and worst models in the natural condition.}
    \label{fig:predictions-natural}
\end{figure}

\subsection{Texturized stimuli}

\begin{figure}[H]
    \centering
    \includegraphics[width=0.6\linewidth]{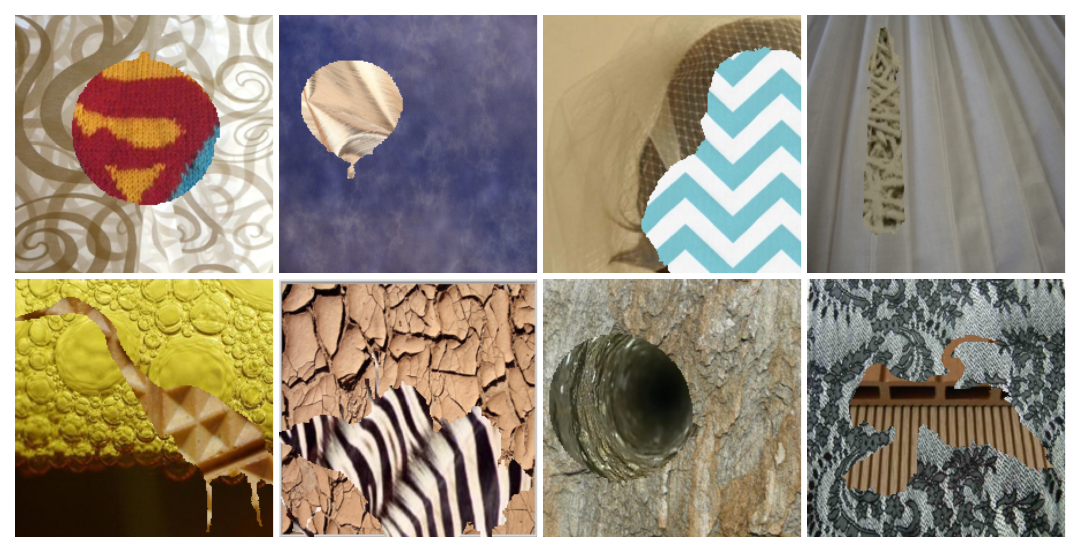}
    \caption{Example images from the texturized test set. The same segmentation masks are used as in the natural condition, but filled with random textures or colors.}
    \label{fig:examples-texturized}
\end{figure}

\begin{table}[H]
  \begin{table}[H]
\centering
\small
\caption{Probe performances on the texturized test set, averaged over all six variants (texture and color variants of the original, reversed, and split stimuli).}
\label{tab:performance-texturized}
\vspace{7pt}
\begin{tabular}{l|rrrrr|rrrrr}
 & \multicolumn{5}{c|}{Trained on natural} & \multicolumn{5}{c}{Trained on texturized} \\
Model & ACC $\downarrow$ & BCE & IG & IGE & mIoU & ACC & BCE & IG & IGE & mIoU \\
\hline
BEiT-3-L & 0.986 & 0.044 & 0.367 & 0.893 & 0.963 & 0.997 & 0.011 & 0.400 & 0.973 & 0.990 \\
EVA-02-L & 0.975 & 0.070 & 0.350 & 0.833 & 0.941 & 0.992 & 0.023 & 0.397 & 0.945 & 0.979 \\
PE-B & 0.968 & 0.093 & 0.361 & 0.796 & 0.933 & 0.991 & 0.025 & 0.429 & 0.945 & 0.980 \\
DINOv3-L & 0.967 & 0.092 & 0.328 & 0.782 & 0.925 & 0.995 & 0.016 & 0.404 & 0.962 & 0.985 \\
CLIP-L & 0.966 & 0.093 & 0.327 & 0.779 & 0.923 & 0.987 & 0.039 & 0.381 & 0.906 & 0.969 \\
PE-L & 0.965 & 0.103 & 0.351 & 0.773 & 0.925 & 0.993 & 0.022 & 0.432 & 0.952 & 0.984 \\
BEiT-3-B & 0.963 & 0.097 & 0.314 & 0.764 & 0.914 & 0.995 & 0.017 & 0.395 & 0.960 & 0.984 \\
CLIP-B & 0.960 & 0.118 & 0.293 & 0.712 & 0.906 & 0.983 & 0.054 & 0.357 & 0.868 & 0.958 \\
MAE-B & 0.959 & 0.113 & 0.298 & 0.725 & 0.914 & 0.987 & 0.037 & 0.375 & 0.911 & 0.968 \\
DINOv2-L & 0.958 & 0.116 & 0.342 & 0.746 & 0.916 & 0.990 & 0.030 & 0.428 & 0.935 & 0.976 \\
SigLIP-2-L & 0.957 & 0.115 & 0.304 & 0.725 & 0.913 & 0.988 & 0.039 & 0.381 & 0.908 & 0.970 \\
DINOv3-B & 0.957 & 0.121 & 0.298 & 0.710 & 0.907 & 0.991 & 0.025 & 0.395 & 0.941 & 0.978 \\
BEiT-L & 0.954 & 0.119 & 0.292 & 0.711 & 0.897 & 0.992 & 0.023 & 0.389 & 0.944 & 0.980 \\
EVA-02-B & 0.951 & 0.122 & 0.298 & 0.709 & 0.894 & 0.987 & 0.037 & 0.384 & 0.913 & 0.968 \\
MAE-L & 0.947 & 0.131 & 0.281 & 0.683 & 0.890 & 0.992 & 0.025 & 0.387 & 0.940 & 0.980 \\
DeiT-3-L & 0.946 & 0.152 & 0.260 & 0.632 & 0.887 & 0.995 & 0.015 & 0.396 & 0.963 & 0.987 \\
DeiT-3-B & 0.940 & 0.162 & 0.249 & 0.606 & 0.869 & 0.993 & 0.023 & 0.389 & 0.945 & 0.981 \\
DINO-B & 0.939 & 0.149 & 0.263 & 0.639 & 0.879 & 0.988 & 0.034 & 0.377 & 0.917 & 0.970 \\
SigLIP-2-B & 0.936 & 0.182 & 0.229 & 0.557 & 0.863 & 0.976 & 0.073 & 0.338 & 0.822 & 0.936 \\
DINOv2-B & 0.933 & 0.164 & 0.294 & 0.642 & 0.867 & 0.983 & 0.048 & 0.410 & 0.895 & 0.960 \\
BEiT-B & 0.928 & 0.176 & 0.236 & 0.573 & 0.840 & 0.987 & 0.039 & 0.372 & 0.905 & 0.966 \\
AugReg-L & 0.909 & 0.244 & 0.167 & 0.406 & 0.806 & 0.976 & 0.071 & 0.340 & 0.828 & 0.939 \\
AugReg-B & 0.898 & 0.277 & 0.134 & 0.326 & 0.776 & 0.972 & 0.078 & 0.334 & 0.811 & 0.928 \\
FlexiViT-L & 0.895 & 0.256 & 0.159 & 0.383 & 0.771 & 0.983 & 0.053 & 0.362 & 0.872 & 0.958 \\
FlexiViT-B & 0.858 & 0.350 & 0.066 & 0.159 & 0.683 & 0.964 & 0.104 & 0.312 & 0.751 & 0.913 \\
\hline
Spatial prior & 0.798 & 0.433 & 0.000 & 0.000 & 0.571 & 0.798 & 0.433 & 0.000 & 0.000 & 0.571 \\
Area & 0.944 & -- & -- & -- & -- & 0.944 & -- & -- & -- & -- \\
Center overlap & 0.715 & -- & -- & -- & -- & 0.715 & -- & -- & -- & -- \\
Offset & 0.901 & -- & -- & -- & -- & 0.901 & -- & -- & -- & -- \\
\end{tabular}
\end{table}

\end{table}

\begin{figure}[H]
    \centering
    \includegraphics[width=0.9\linewidth]{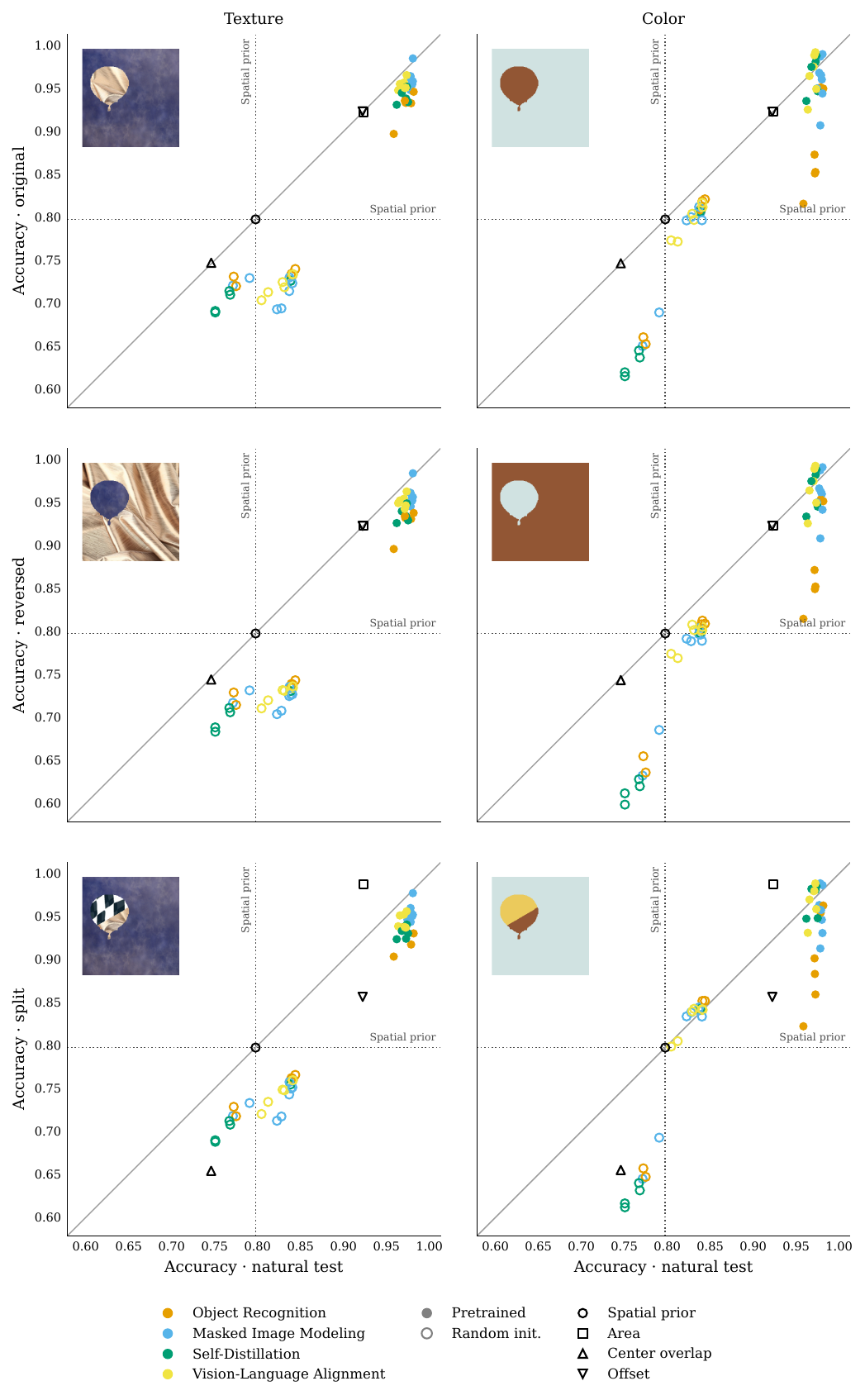}
    \caption{Comparison of the probe performances on the i.i.d. test set of natural images (x-axis) and applied zero-shot to texturized stimulus variants (y-axis).}
    \label{fig:natural-vs-texturized-variants}
\end{figure}

\clearpage
\begin{figure}[H]
    \centering
    \includegraphics[width=0.85\linewidth]{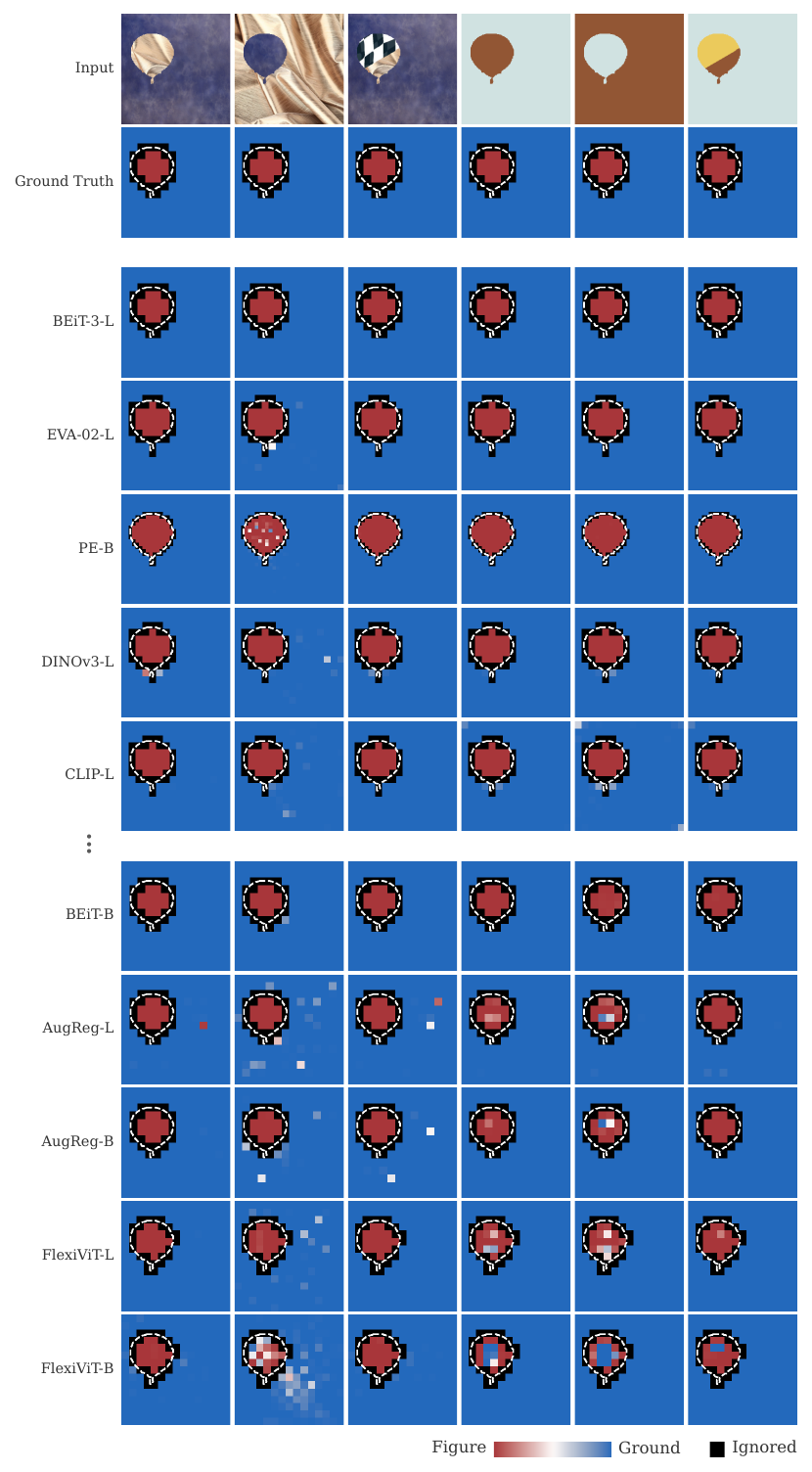}
    \caption{Probe predictions for the best and worst models in the texturized condition.}
    \label{fig:predictions-texturized}
\end{figure}

\clearpage
\begin{figure}[H]
    \centering
    \includegraphics[width=0.9\linewidth]{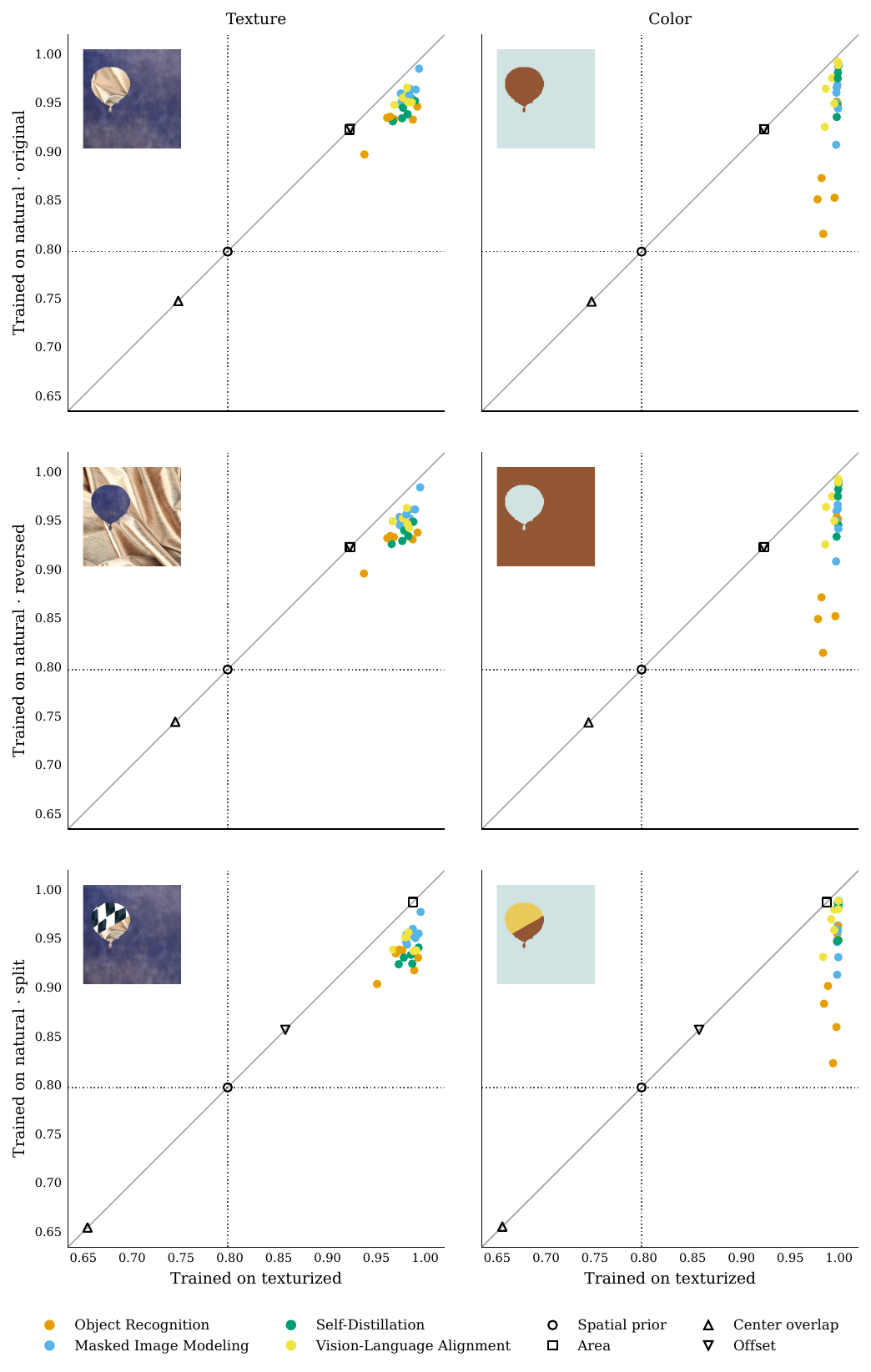}
    \caption{Performance comparison of probes directly trained on the texturized images (x-axis) with zero-shot probes trained on natural images (y-axis).}
    \label{fig:direct-training-texturized-variants}
\end{figure}

\subsection{Surroundedness}

\begin{figure}[H]
    \centering
    \includegraphics[width=0.6\linewidth]{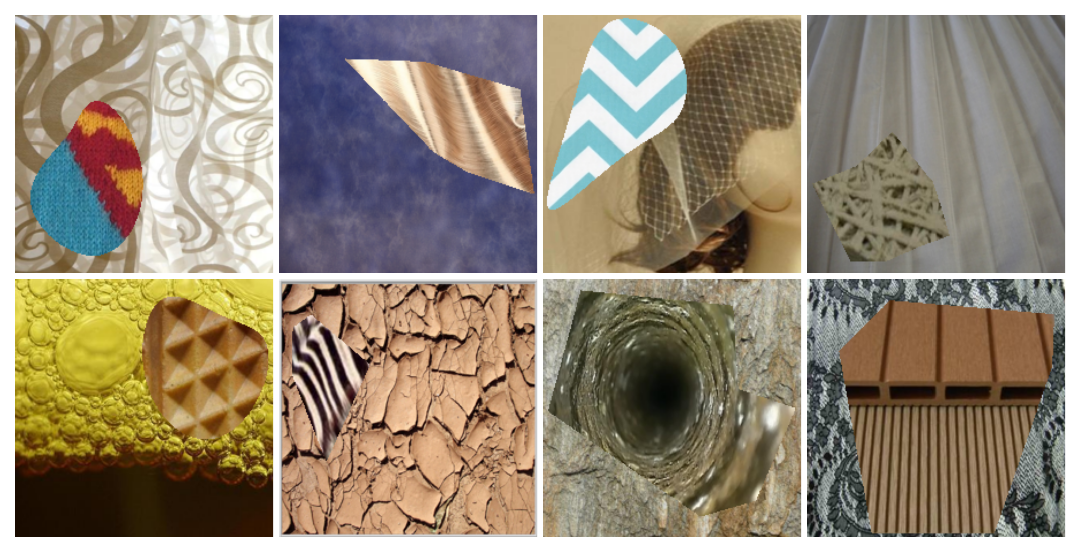}
    \caption{Example images from the surroundedness test set. Each image contains a random shape from the idSprites dataset \cite{dziadzio2024infinite}, surroundend by the background on all sides.}
    \label{fig:examples-surroundedness}
\end{figure}

\begin{table}[H]
  \begin{table}[H]
\centering
\small
\caption{Probe performances on the surroundedness test set, averaged over all six variants (texture and color variants of the original, reversed, and split stimuli).}
\label{tab:performance-surroundedness}
\vspace{7pt}
\begin{tabular}{l|rrrrr|rrrrr}
 & \multicolumn{5}{c|}{Trained on natural} & \multicolumn{5}{c}{Trained on surroundedness} \\
Model & ACC $\downarrow$ & BCE & IG & IGE & mIoU & ACC & BCE & IG & IGE & mIoU \\
\hline
BEiT-3-L & 0.986 & 0.051 & 0.439 & 0.897 & 0.970 & 0.999 & 0.004 & 0.419 & 0.990 & 0.996 \\
EVA-02-L & 0.981 & 0.066 & 0.430 & 0.867 & 0.958 & 0.995 & 0.014 & 0.418 & 0.967 & 0.988 \\
PE-B & 0.975 & 0.073 & 0.447 & 0.859 & 0.952 & 0.996 & 0.014 & 0.453 & 0.970 & 0.990 \\
DINOv3-L & 0.968 & 0.096 & 0.400 & 0.806 & 0.935 & 0.997 & 0.009 & 0.424 & 0.979 & 0.991 \\
PE-L & 0.965 & 0.095 & 0.425 & 0.817 & 0.931 & 0.997 & 0.009 & 0.458 & 0.981 & 0.993 \\
DINOv3-B & 0.963 & 0.110 & 0.386 & 0.779 & 0.928 & 0.995 & 0.014 & 0.419 & 0.967 & 0.988 \\
CLIP-L & 0.961 & 0.101 & 0.396 & 0.797 & 0.922 & 0.992 & 0.032 & 0.401 & 0.926 & 0.980 \\
DINOv2-L & 0.961 & 0.118 & 0.405 & 0.774 & 0.924 & 0.992 & 0.024 & 0.447 & 0.949 & 0.980 \\
BEiT-L & 0.956 & 0.110 & 0.380 & 0.775 & 0.916 & 0.996 & 0.011 & 0.412 & 0.973 & 0.989 \\
DINO-B & 0.951 & 0.126 & 0.364 & 0.743 & 0.912 & 0.992 & 0.024 & 0.400 & 0.944 & 0.979 \\
CLIP-B & 0.950 & 0.151 & 0.338 & 0.691 & 0.900 & 0.989 & 0.051 & 0.372 & 0.880 & 0.973 \\
EVA-02-B & 0.950 & 0.118 & 0.379 & 0.762 & 0.904 & 0.991 & 0.027 & 0.405 & 0.937 & 0.977 \\
MAE-B & 0.950 & 0.122 & 0.368 & 0.752 & 0.911 & 0.992 & 0.023 & 0.400 & 0.945 & 0.981 \\
BEiT-3-B & 0.948 & 0.120 & 0.370 & 0.755 & 0.903 & 0.997 & 0.010 & 0.413 & 0.976 & 0.989 \\
DeiT-3-B & 0.943 & 0.141 & 0.349 & 0.712 & 0.887 & 0.996 & 0.013 & 0.410 & 0.969 & 0.988 \\
DeiT-3-L & 0.940 & 0.136 & 0.354 & 0.723 & 0.891 & 0.998 & 0.007 & 0.416 & 0.983 & 0.994 \\
MAE-L & 0.931 & 0.150 & 0.340 & 0.694 & 0.881 & 0.996 & 0.014 & 0.409 & 0.967 & 0.989 \\
SigLIP-2-L & 0.919 & 0.166 & 0.330 & 0.665 & 0.865 & 0.993 & 0.026 & 0.407 & 0.941 & 0.982 \\
DINOv2-B & 0.918 & 0.179 & 0.345 & 0.658 & 0.852 & 0.988 & 0.036 & 0.435 & 0.923 & 0.970 \\
SigLIP-2-B & 0.907 & 0.263 & 0.227 & 0.463 & 0.833 & 0.986 & 0.051 & 0.373 & 0.880 & 0.961 \\
BEiT-B & 0.899 & 0.211 & 0.279 & 0.569 & 0.822 & 0.991 & 0.027 & 0.396 & 0.935 & 0.977 \\
AugReg-L & 0.867 & 0.384 & 0.105 & 0.215 & 0.767 & 0.984 & 0.052 & 0.371 & 0.876 & 0.958 \\
AugReg-B & 0.865 & 0.394 & 0.095 & 0.195 & 0.764 & 0.977 & 0.069 & 0.355 & 0.837 & 0.941 \\
FlexiViT-L & 0.827 & 0.414 & 0.080 & 0.162 & 0.701 & 0.988 & 0.044 & 0.384 & 0.896 & 0.969 \\
FlexiViT-B & 0.787 & 0.504 & -0.010 & -0.020 & 0.616 & 0.967 & 0.109 & 0.319 & 0.745 & 0.919 \\
\hline
Spatial prior & 0.721 & 0.506 & 0.000 & 0.000 & 0.479 & 0.760 & 0.446 & 0.000 & 0.000 & 0.582 \\
Area & 0.660 & -- & -- & -- & -- & 0.660 & -- & -- & -- & -- \\
Center overlap & 0.535 & -- & -- & -- & -- & 0.535 & -- & -- & -- & -- \\
Offset & 0.598 & -- & -- & -- & -- & 0.598 & -- & -- & -- & -- \\
\end{tabular}
\end{table}

\end{table}

\begin{figure}[!ht]
    \centering
    \includegraphics[width=0.9\linewidth]{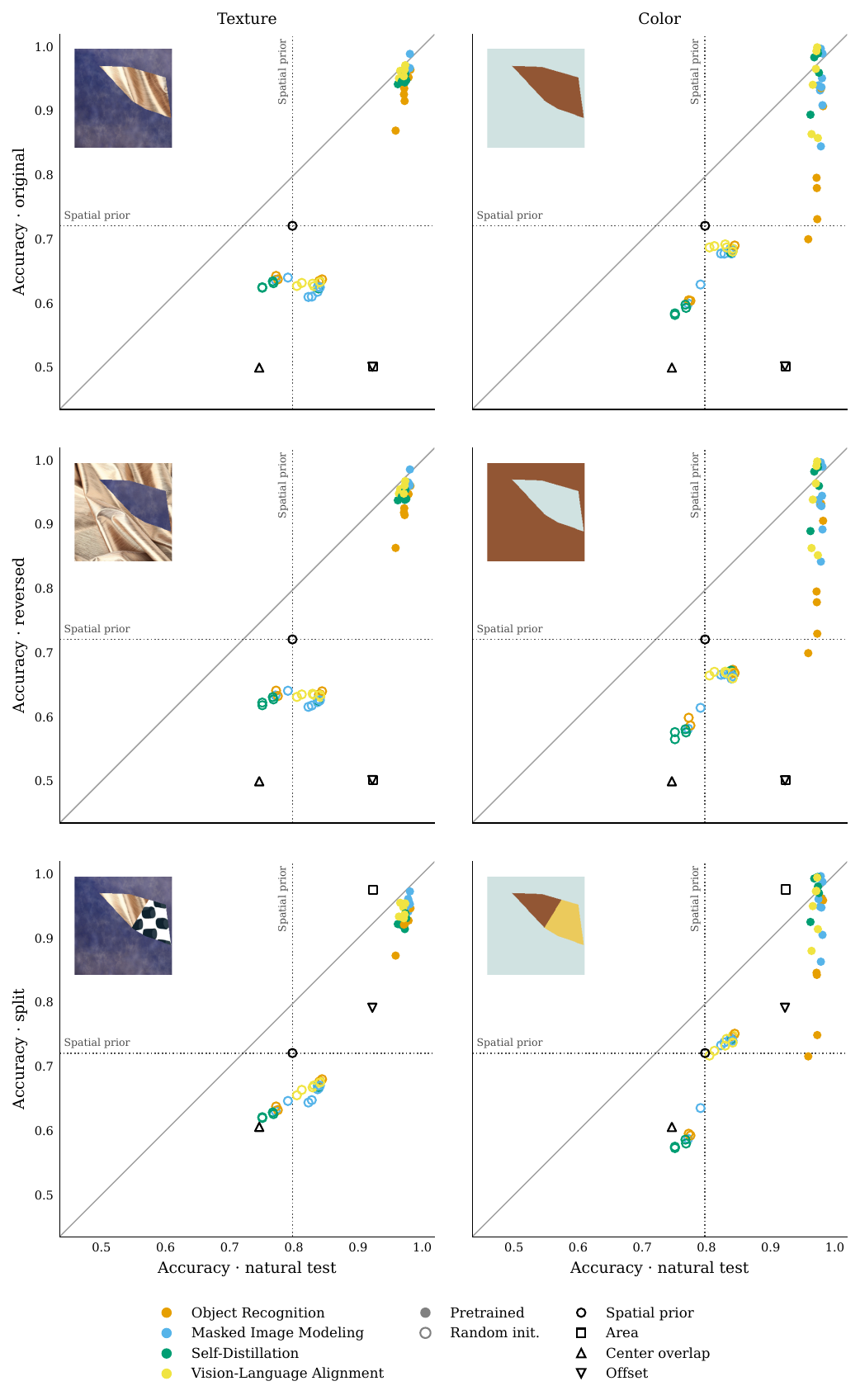}
    \caption{Comparison of the probe performances on the i.i.d. test set of natural images (x-axis) and applied zero-shot to the stimuli isolating surroundedness (y-axis).}
    \label{fig:natural-vs-surroundedness-variants}
\end{figure}

\clearpage
\begin{figure}[H]
    \centering
    \includegraphics[width=0.85\linewidth]{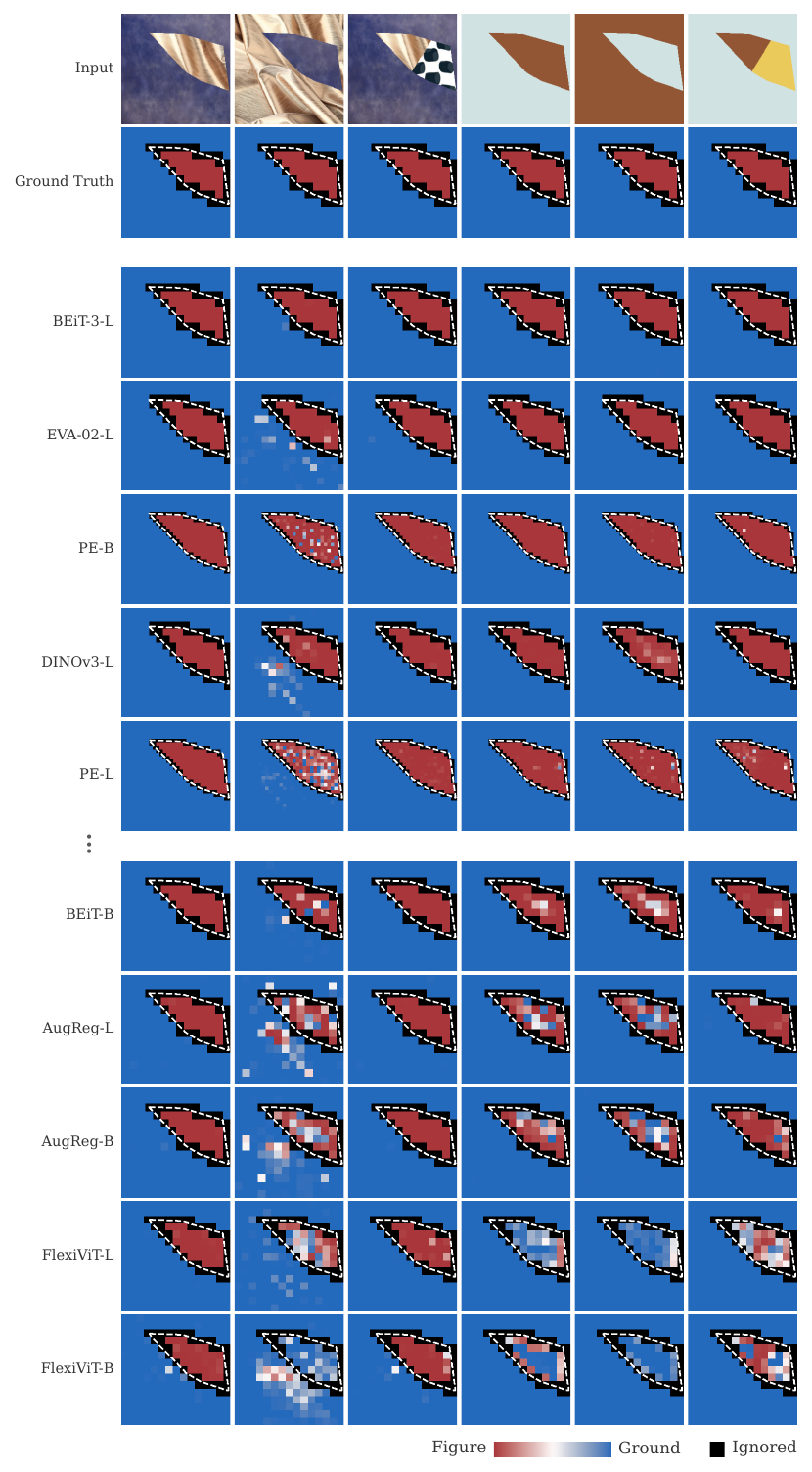}
    \caption{Probe predictions for the best and worst models in the surroundedness condition.}
    \label{fig:predictions-surroundedness}
\end{figure}

\clearpage
\begin{figure}[H]
    \centering
    \includegraphics[width=0.9\linewidth]{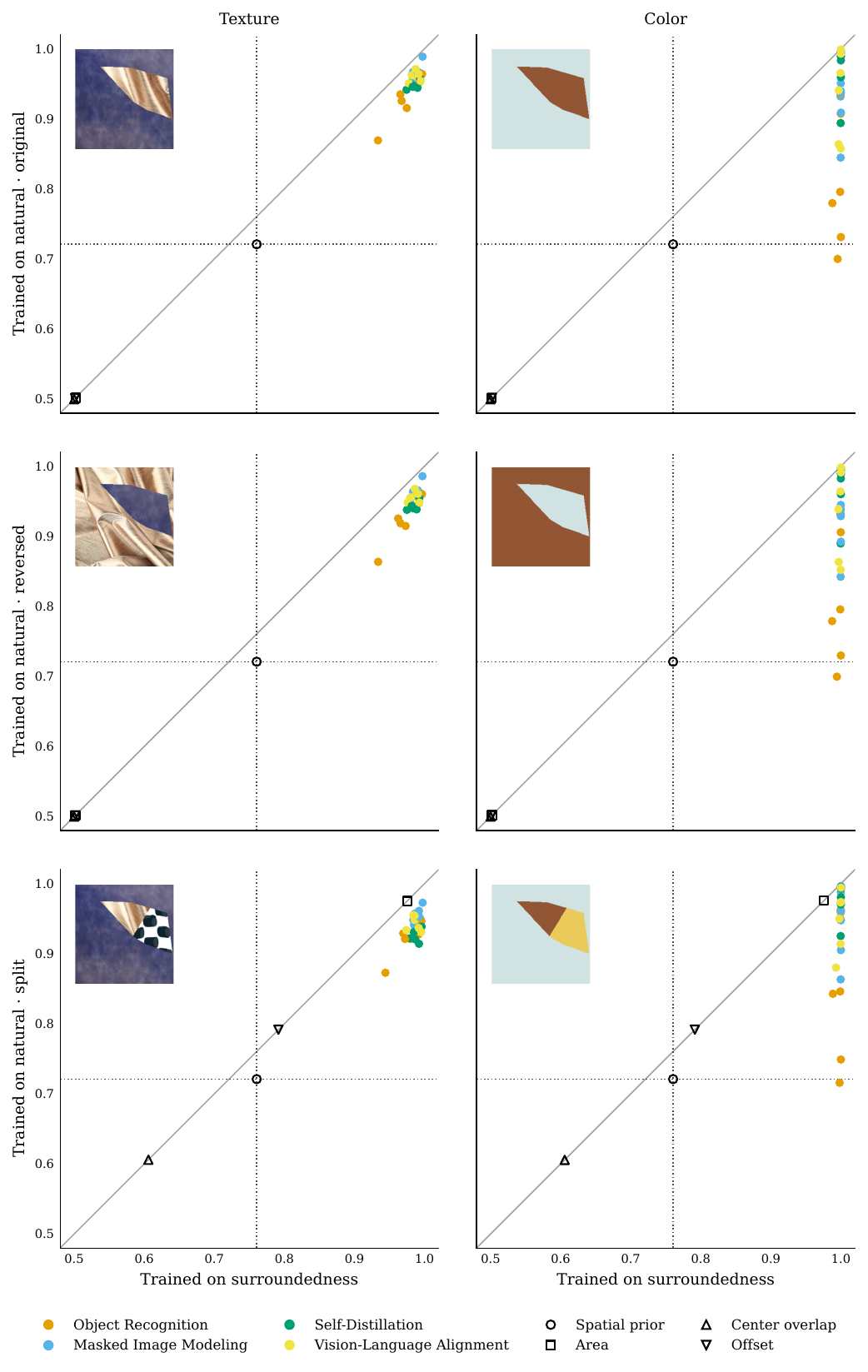}
    \caption{Performance comparison of probes directly trained on the surroundedness stimuli (x-axis) with zero-shot probes trained on natural images (y-axis).}
    \label{fig:direct-training-urroundedness-variants}
\end{figure}

\subsection{Convexity}

\begin{figure}[H]
    \centering
    \includegraphics[width=0.6\linewidth]{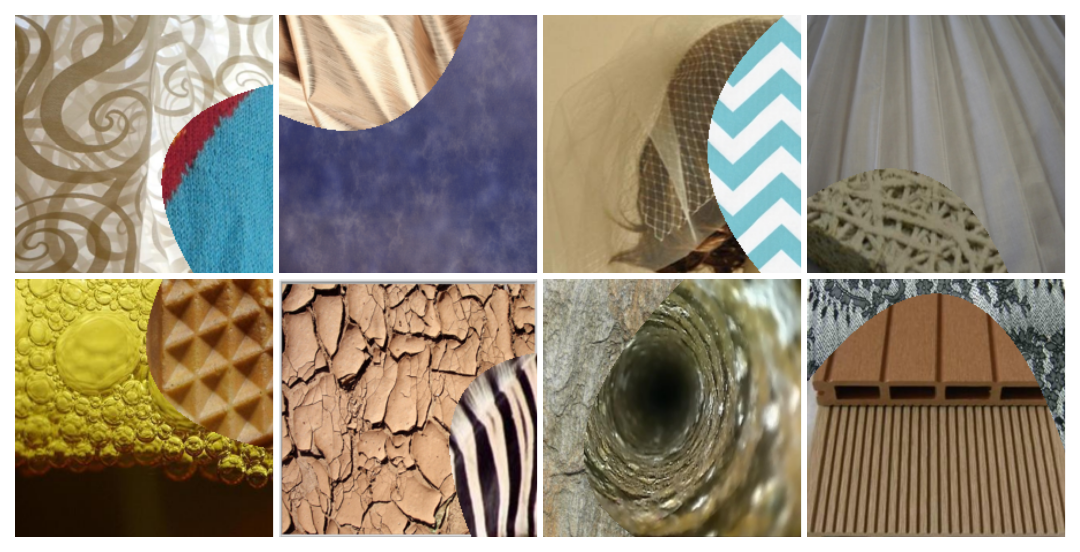}
    \caption{Example images from the convexity condition. Each image contains two regions separated by a random parabolic arc, filled with random textures or colors.}
    \label{fig:examples-convexity}
\end{figure}

\begin{table}[H]
  \begin{table}[H]
\centering
\small
\caption{Probe performances on the convexity test sets, averaged over all six variants (texture and color variants of the original, reversed, and split stimuli).}
\label{tab:performance-convexity}
\vspace{7pt}
\begin{tabular}{l|rrrrr|rrrrr}
 & \multicolumn{5}{c|}{Trained on natural} & \multicolumn{5}{c}{Trained on convexity} \\
Model & ACC $\downarrow$ & BCE & IG & IGE & mIoU & ACC & BCE & IG & IGE & mIoU \\
\hline
BEiT-3-L & 0.730 & 0.683 & 0.359 & 0.344 & 0.593 & 0.977 & 0.072 & 0.604 & 0.894 & 0.952 \\
EVA-02-L & 0.703 & 0.711 & 0.313 & 0.305 & 0.547 & 0.953 & 0.131 & 0.545 & 0.806 & 0.909 \\
PE-B & 0.699 & 0.945 & 0.024 & 0.025 & 0.580 & 0.952 & 0.127 & 0.549 & 0.812 & 0.914 \\
CLIP-L & 0.685 & 0.810 & 0.214 & 0.209 & 0.533 & 0.931 & 0.178 & 0.498 & 0.737 & 0.855 \\
DINOv2-L & 0.677 & 0.732 & 0.229 & 0.238 & 0.516 & 0.944 & 0.140 & 0.536 & 0.793 & 0.897 \\
CLIP-B & 0.675 & 0.728 & 0.314 & 0.301 & 0.509 & 0.915 & 0.234 & 0.441 & 0.653 & 0.836 \\
DINO-B & 0.672 & 0.881 & 0.161 & 0.154 & 0.532 & 0.964 & 0.094 & 0.581 & 0.860 & 0.931 \\
BEiT-L & 0.659 & 0.988 & 0.053 & 0.051 & 0.509 & 0.963 & 0.099 & 0.576 & 0.853 & 0.926 \\
PE-L & 0.652 & 1.087 & -0.118 & -0.122 & 0.525 & 0.962 & 0.105 & 0.571 & 0.844 & 0.930 \\
MAE-B & 0.644 & 0.991 & 0.051 & 0.049 & 0.480 & 0.919 & 0.199 & 0.477 & 0.706 & 0.864 \\
DINOv2-B & 0.642 & 0.950 & 0.010 & 0.011 & 0.462 & 0.927 & 0.177 & 0.499 & 0.738 & 0.870 \\
DINOv3-L & 0.640 & 1.085 & -0.060 & -0.059 & 0.459 & 0.968 & 0.088 & 0.588 & 0.870 & 0.935 \\
BEiT-B & 0.632 & 1.137 & -0.095 & -0.092 & 0.460 & 0.935 & 0.173 & 0.503 & 0.744 & 0.879 \\
EVA-02-B & 0.631 & 1.089 & -0.065 & -0.064 & 0.445 & 0.924 & 0.199 & 0.476 & 0.705 & 0.856 \\
BEiT-3-B & 0.631 & 1.081 & -0.039 & -0.038 & 0.447 & 0.963 & 0.111 & 0.565 & 0.836 & 0.925 \\
SigLIP-2-L & 0.627 & 1.034 & -0.009 & -0.008 & 0.448 & 0.925 & 0.205 & 0.470 & 0.696 & 0.854 \\
AugReg-L & 0.626 & 1.233 & -0.191 & -0.184 & 0.453 & 0.889 & 0.259 & 0.416 & 0.616 & 0.771 \\
MAE-L & 0.622 & 1.133 & -0.091 & -0.088 & 0.447 & 0.939 & 0.155 & 0.521 & 0.771 & 0.895 \\
DeiT-3-L & 0.618 & 1.413 & -0.372 & -0.357 & 0.418 & 0.972 & 0.081 & 0.594 & 0.880 & 0.939 \\
SigLIP-2-B & 0.612 & 1.121 & -0.079 & -0.076 & 0.414 & 0.879 & 0.283 & 0.392 & 0.580 & 0.761 \\
AugReg-B & 0.608 & 1.302 & -0.261 & -0.250 & 0.419 & 0.893 & 0.251 & 0.424 & 0.628 & 0.777 \\
DINOv3-B & 0.607 & 1.204 & -0.178 & -0.174 & 0.428 & 0.958 & 0.110 & 0.566 & 0.837 & 0.921 \\
DeiT-3-B & 0.604 & 1.324 & -0.283 & -0.271 & 0.391 & 0.958 & 0.111 & 0.564 & 0.835 & 0.909 \\
FlexiViT-L & 0.597 & 1.386 & -0.354 & -0.342 & 0.399 & 0.947 & 0.167 & 0.509 & 0.753 & 0.890 \\
FlexiViT-B & 0.555 & 1.354 & -0.321 & -0.311 & 0.338 & 0.902 & 0.273 & 0.402 & 0.596 & 0.811 \\
\hline
Spatial prior & 0.531 & 1.003 & 0.000 & 0.000 & 0.308 & 0.582 & 0.676 & 0.000 & 0.000 & 0.397 \\
Area & 0.548 & -- & -- & -- & -- & 0.548 & -- & -- & -- & -- \\
Center overlap & 0.550 & -- & -- & -- & -- & 0.550 & -- & -- & -- & -- \\
Offset & 0.506 & -- & -- & -- & -- & 0.506 & -- & -- & -- & -- \\
\end{tabular}
\end{table}

\end{table}

\begin{figure}[!ht]
    \centering
    \includegraphics[width=0.9\linewidth]{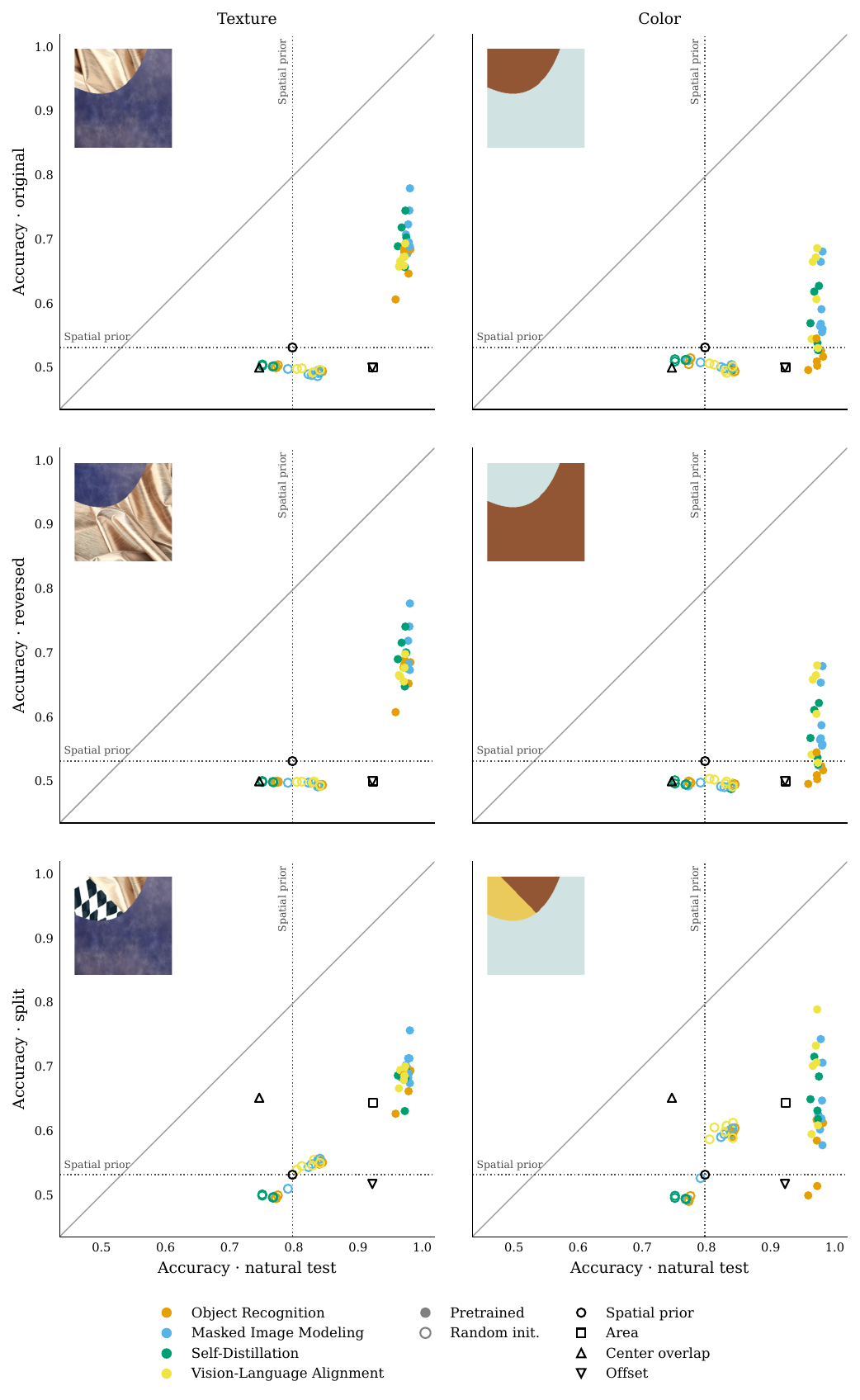}
    \caption{Comparison of the probe performances on the i.i.d. test set of natural images (x-axis) and applied zero-shot to the stimuli isolating convexity (y-axis).}
    \label{fig:natural-vs-convexity-variants}
\end{figure}

\clearpage
\begin{figure}[H]
    \centering
    \includegraphics[width=0.85\linewidth]{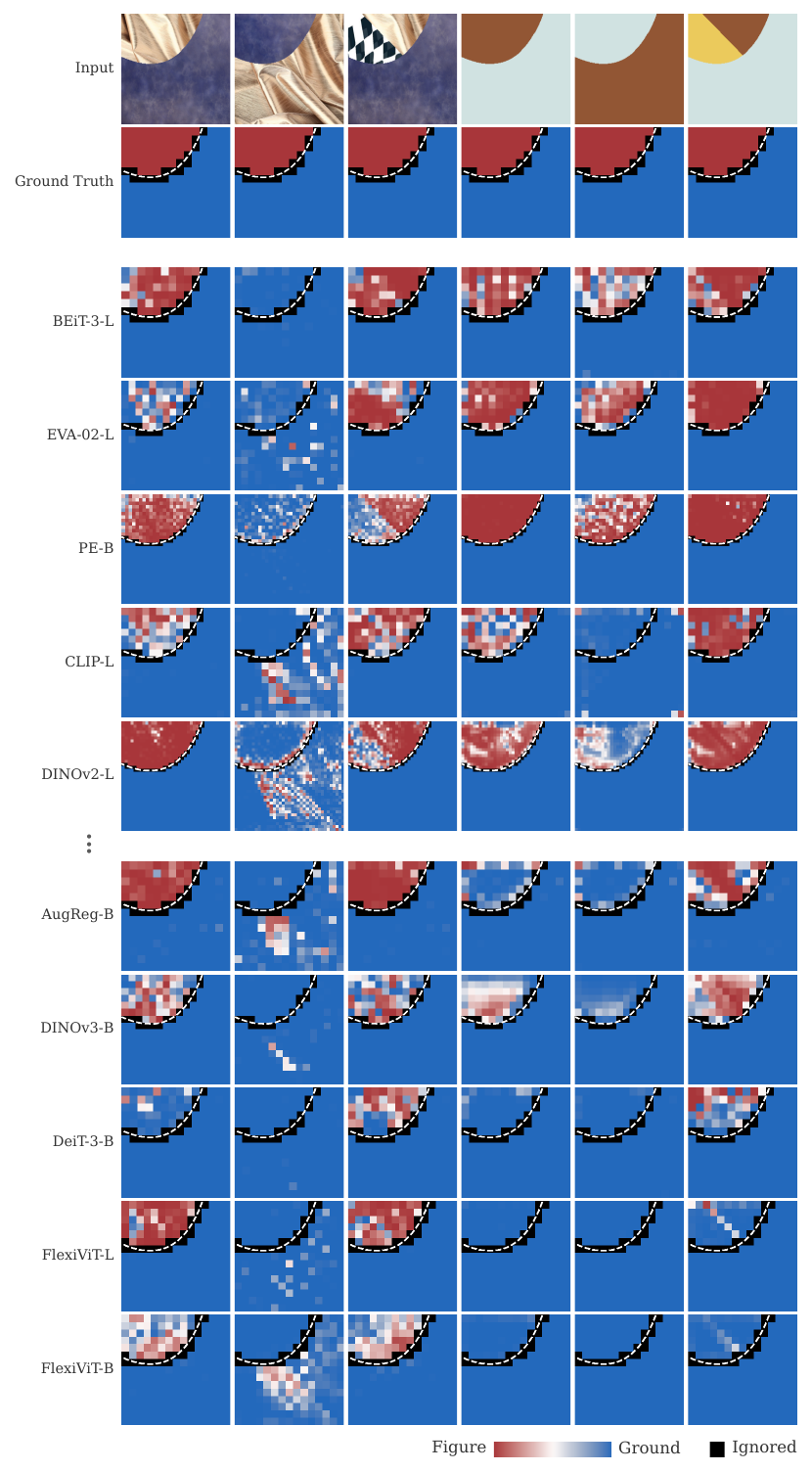}
    \caption{Probe predictions for the best and worst models in the convexity condition.}
    \label{fig:predictions-convexity}
\end{figure}
\clearpage

\clearpage
\begin{figure}[H]
    \centering
    \includegraphics[width=0.9\linewidth]{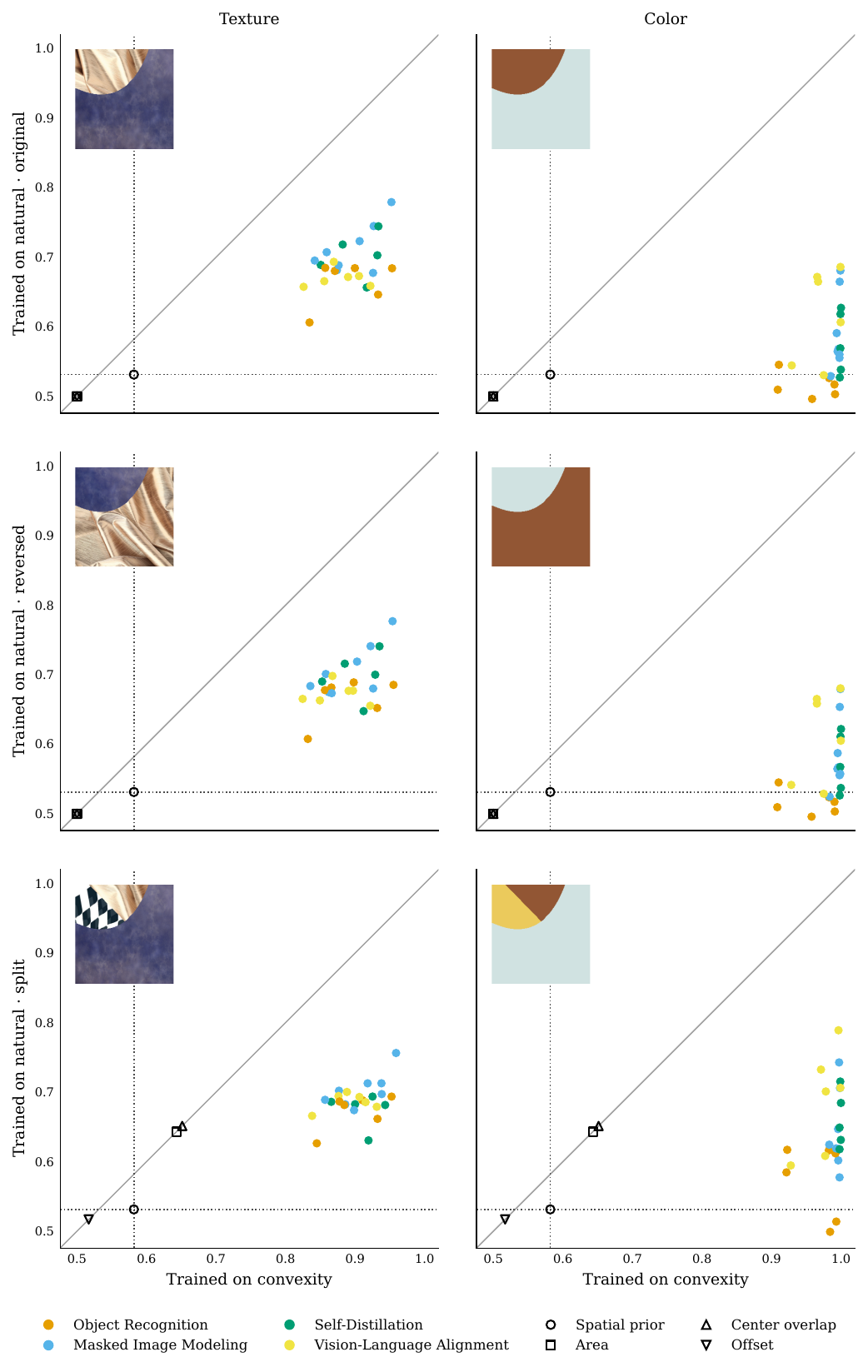}
    \caption{Performance comparison of probes directly trained on the convexity stimuli (x-axis) with zero-shot probes trained on natural images (y-axis).}
    \label{fig:direct-training-convexity-variants}
\end{figure}

\subsection{Symmetry}

\begin{figure}[H]
    \centering
    \includegraphics[width=0.6\linewidth]{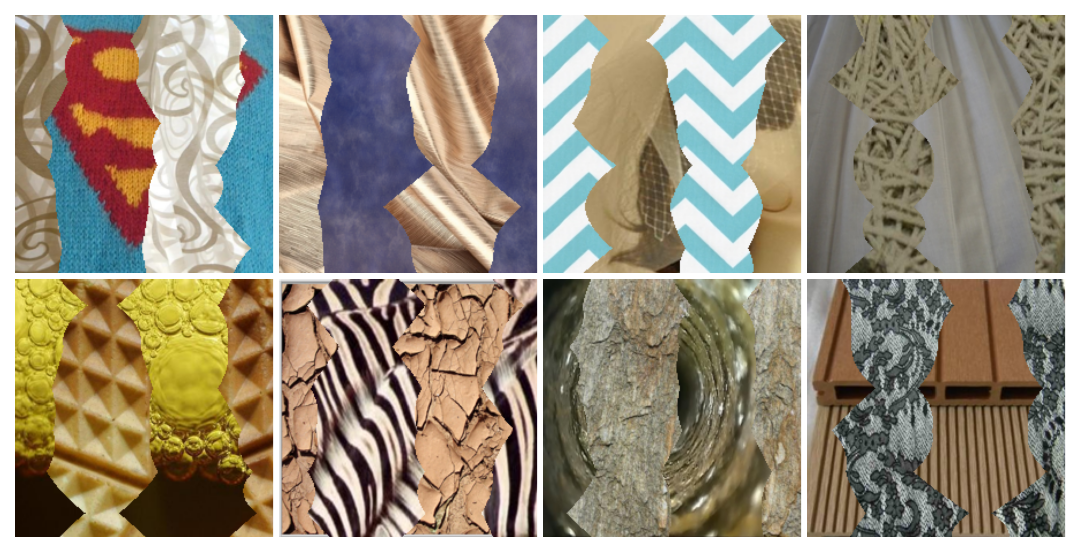}
    \caption{Example images from the symmetry condition. Each image is divided into columns, with one of the columns having mirror symmetric contours.}
    \label{fig:examples-symmetry}
\end{figure}

\begin{table}[H]
  \begin{table}[H]
\centering
\small
\caption{Probe performances on the symmetry test sets, averaged over all four variants (texture and color variants of the original and reversed stimuli).}
\label{tab:performance-symmetry}
\vspace{7pt}
\begin{tabular}{l|rrrrr|rrrrr}
 & \multicolumn{5}{c|}{Trained on natural} & \multicolumn{5}{c}{Trained on symmetry} \\
Model & ACC $\downarrow$ & BCE & IG & IGE & mIoU & ACC & BCE & IG & IGE & mIoU \\
\hline
PE-B & 0.567 & 1.781 & -0.700 & -0.648 & 0.425 & 0.723 & 0.537 & 0.157 & 0.226 & 0.625 \\
DINOv2-L & 0.552 & 1.217 & -0.147 & -0.137 & 0.394 & 0.818 & 0.372 & 0.321 & 0.463 & 0.739 \\
PE-L & 0.543 & 1.747 & -0.667 & -0.618 & 0.359 & 0.776 & 0.455 & 0.239 & 0.344 & 0.697 \\
DINOv3-L & 0.532 & 1.720 & -0.569 & -0.494 & 0.351 & 0.800 & 0.394 & 0.300 & 0.432 & 0.719 \\
EVA-02-L & 0.531 & 1.573 & -0.423 & -0.368 & 0.368 & 0.745 & 0.547 & 0.147 & 0.211 & 0.640 \\
CLIP-L & 0.528 & 1.448 & -0.299 & -0.260 & 0.353 & 0.726 & 0.537 & 0.156 & 0.225 & 0.606 \\
CLIP-B & 0.526 & 1.253 & -0.080 & -0.068 & 0.340 & 0.666 & 0.605 & 0.089 & 0.128 & 0.522 \\
DINOv3-B & 0.525 & 1.766 & -0.615 & -0.534 & 0.328 & 0.772 & 0.447 & 0.246 & 0.355 & 0.687 \\
DINOv2-B & 0.521 & 1.459 & -0.389 & -0.363 & 0.345 & 0.745 & 0.497 & 0.197 & 0.284 & 0.639 \\
SigLIP-2-L & 0.521 & 1.566 & -0.414 & -0.360 & 0.326 & 0.746 & 0.529 & 0.165 & 0.237 & 0.632 \\
DeiT-3-B & 0.518 & 1.876 & -0.703 & -0.600 & 0.317 & 0.653 & 0.609 & 0.085 & 0.122 & 0.504 \\
BEiT-L & 0.514 & 1.611 & -0.438 & -0.374 & 0.326 & 0.792 & 0.403 & 0.291 & 0.420 & 0.714 \\
DeiT-3-L & 0.511 & 2.128 & -0.955 & -0.814 & 0.305 & 0.742 & 0.527 & 0.166 & 0.240 & 0.632 \\
SigLIP-2-B & 0.510 & 1.575 & -0.402 & -0.343 & 0.305 & 0.651 & 0.619 & 0.075 & 0.108 & 0.500 \\
BEiT-B & 0.510 & 1.673 & -0.500 & -0.426 & 0.314 & 0.743 & 0.521 & 0.173 & 0.249 & 0.638 \\
EVA-02-B & 0.509 & 1.911 & -0.762 & -0.663 & 0.315 & 0.693 & 0.586 & 0.107 & 0.155 & 0.559 \\
BEiT-3-B & 0.508 & 2.009 & -0.836 & -0.713 & 0.303 & 0.760 & 0.494 & 0.200 & 0.288 & 0.666 \\
DINO-B & 0.507 & 1.824 & -0.651 & -0.555 & 0.302 & 0.648 & 0.640 & 0.054 & 0.078 & 0.502 \\
FlexiViT-L & 0.506 & 1.976 & -0.814 & -0.700 & 0.290 & 0.680 & 0.612 & 0.082 & 0.118 & 0.538 \\
AugReg-L & 0.505 & 1.760 & -0.587 & -0.501 & 0.304 & 0.659 & 0.606 & 0.087 & 0.126 & 0.509 \\
AugReg-B & 0.505 & 1.756 & -0.583 & -0.497 & 0.297 & 0.607 & 0.655 & 0.039 & 0.056 & 0.448 \\
MAE-B & 0.505 & 2.344 & -1.171 & -0.999 & 0.300 & 0.758 & 0.502 & 0.192 & 0.276 & 0.656 \\
FlexiViT-B & 0.504 & 1.723 & -0.560 & -0.482 & 0.281 & 0.610 & 0.668 & 0.026 & 0.037 & 0.451 \\
MAE-L & 0.502 & 2.371 & -1.198 & -1.021 & 0.296 & 0.794 & 0.443 & 0.250 & 0.361 & 0.710 \\
BEiT-3-L & 0.501 & 1.657 & -0.484 & -0.412 & 0.338 & 0.802 & 0.431 & 0.262 & 0.378 & 0.722 \\
\hline
Spatial prior & 0.501 & 1.124 & 0.000 & 0.000 & 0.306 & 0.476 & 0.694 & 0.000 & 0.000 & 0.437 \\
Area & 0.523 & -- & -- & -- & -- & 0.523 & -- & -- & -- & -- \\
Center overlap & 0.506 & -- & -- & -- & -- & 0.506 & -- & -- & -- & -- \\
Offset & 0.523 & -- & -- & -- & -- & 0.523 & -- & -- & -- & -- \\
\end{tabular}
\end{table}

\end{table}

\begin{figure}[!ht]
    \centering
    \includegraphics[width=0.9\linewidth]{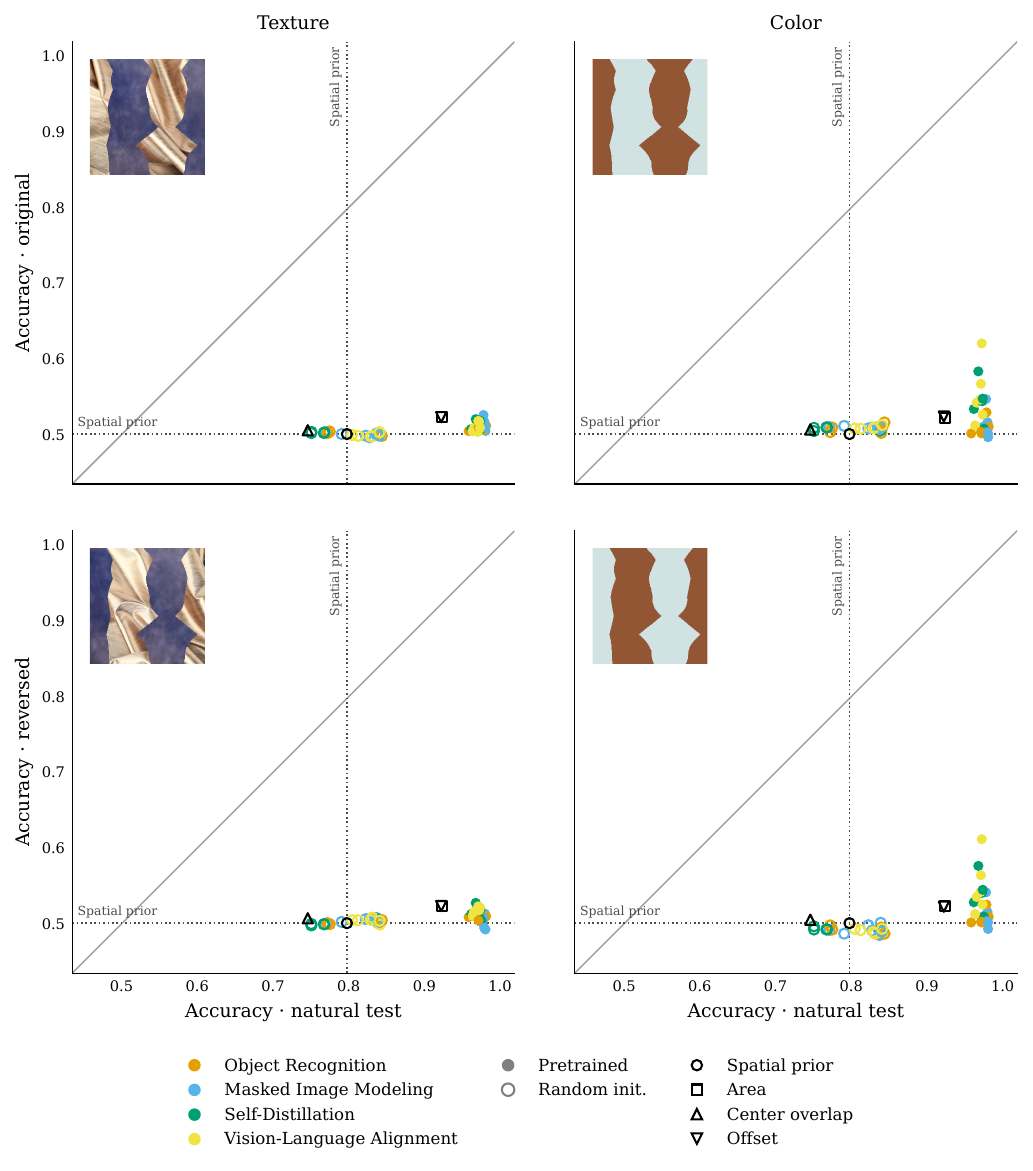}
    \caption{Comparison of the probe performances on the i.i.d. test set of natural images (x-axis) and applied zero-shot to the stimuli isolating symmetry (y-axis).}
    \label{fig:natural-vs-symmetry-variants}
\end{figure}

\clearpage
\begin{figure}[H]
    \centering
    \includegraphics[width=\linewidth]{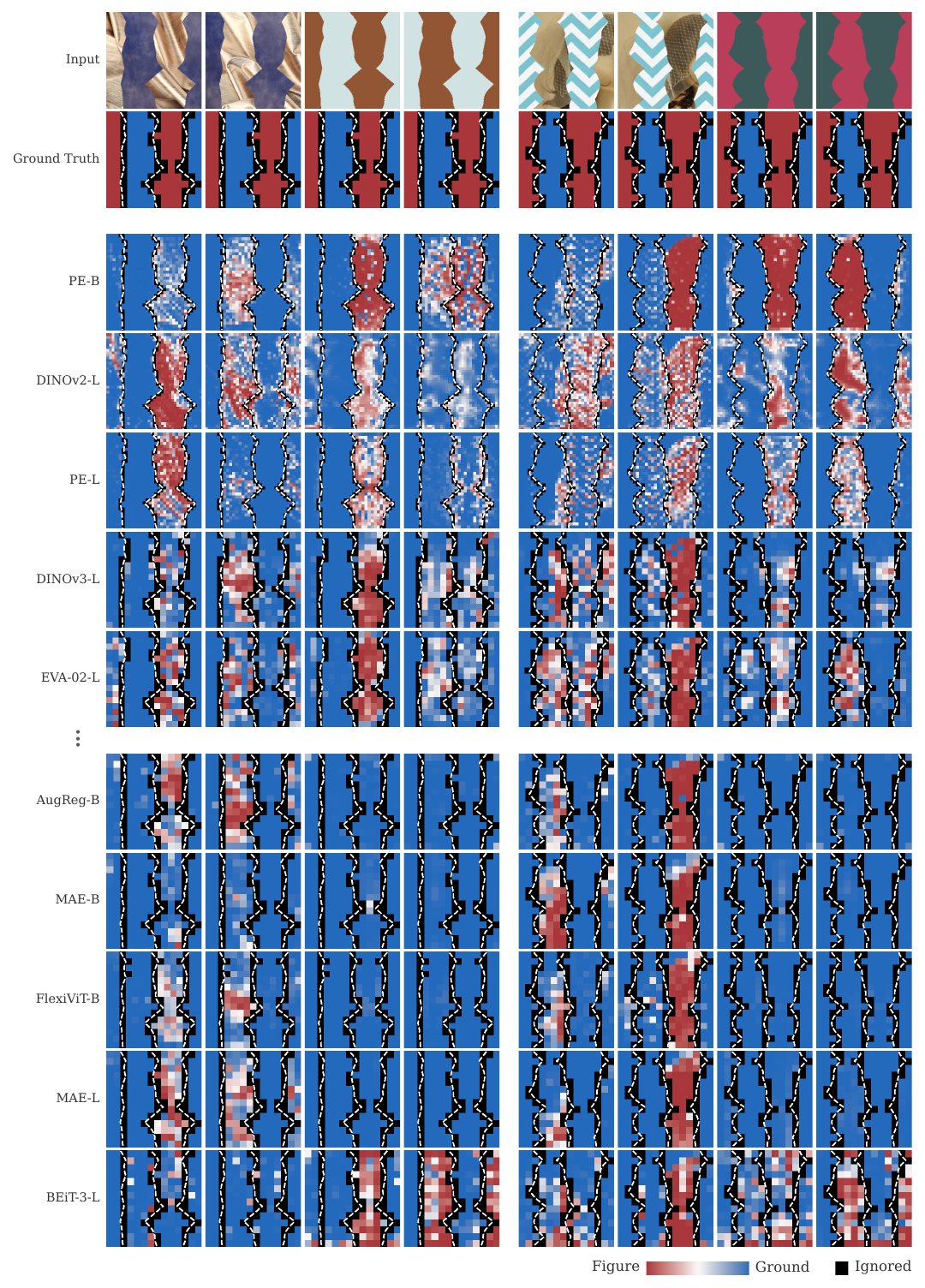}
    \caption{Probe predictions for the best and worst models in the symmetry condition.}
    \label{fig:predictions-symmetry}
\end{figure}

\clearpage
\begin{figure}[H]
    \centering
    \includegraphics[width=0.9\linewidth]{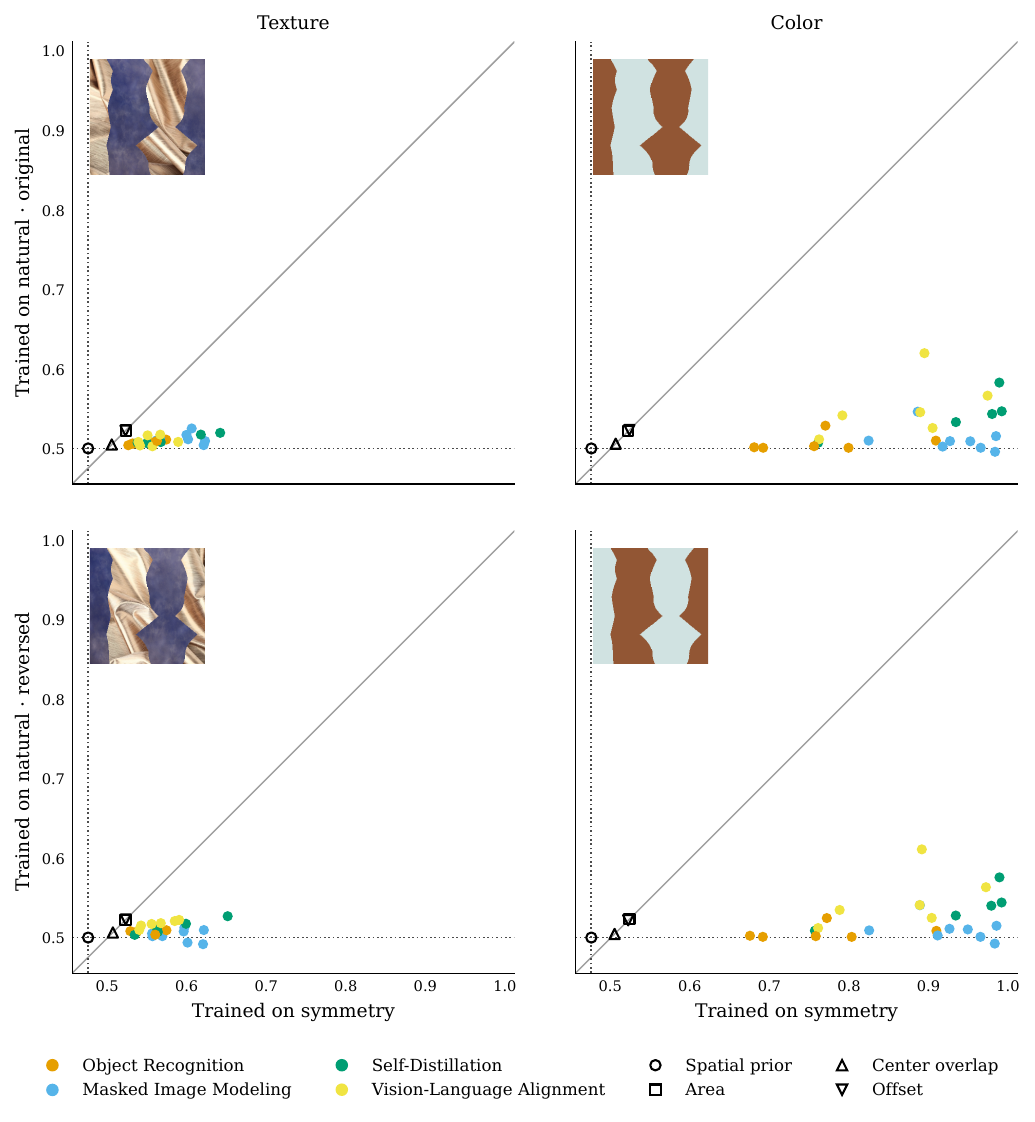}
    \caption{Performance comparison of probes directly trained on the symmetry stimuli (x-axis) with zero-shot probes trained on natural images (y-axis).}
    \label{fig:direct-training-symmetry-variants}
\end{figure}

\clearpage
\subsection{Parallelism}

\begin{figure}[H]
    \centering
    \includegraphics[width=0.6\linewidth]{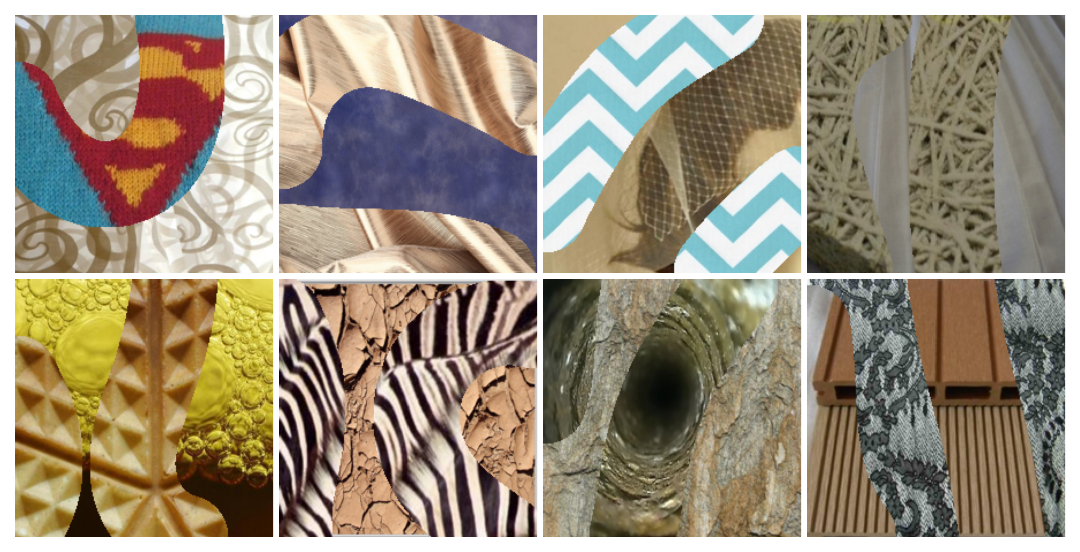}
    \caption{Example images from the parallelism condition, showing random stripes of constant width. Foreground stripes and background are filled with random textures or colors.}
    \label{fig:examples-parallelism}
\end{figure}

\begin{table}[H]
  \begin{table}[H]
\centering
\small
\caption{Probe performances on the parallelism test sets, averaged over all four variants (texture and color variants of the original and reversed stimuli).}
\label{tab:performance-parallelism}
\vspace{7pt}
\begin{tabular}{l|rrrrr|rrrrr}
 & \multicolumn{5}{c|}{Trained on natural} & \multicolumn{5}{c}{Trained on parallelism} \\
Model & ACC $\downarrow$ & BCE & IG & IGE & mIoU & ACC & BCE & IG & IGE & mIoU \\
\hline
DINOv2-L & 0.652 & 0.841 & 0.212 & 0.201 & 0.504 & 0.895 & 0.235 & 0.449 & 0.657 & 0.841 \\
BEiT-3-L & 0.635 & 1.089 & 0.041 & 0.036 & 0.478 & 0.918 & 0.213 & 0.469 & 0.688 & 0.875 \\
PE-L & 0.613 & 1.249 & -0.189 & -0.178 & 0.455 & 0.909 & 0.216 & 0.467 & 0.683 & 0.864 \\
PE-B & 0.611 & 1.279 & -0.216 & -0.203 & 0.465 & 0.863 & 0.315 & 0.368 & 0.539 & 0.803 \\
EVA-02-L & 0.599 & 1.163 & -0.055 & -0.049 & 0.440 & 0.893 & 0.273 & 0.409 & 0.600 & 0.837 \\
DINOv3-L & 0.592 & 1.137 & -0.022 & -0.019 & 0.418 & 0.916 & 0.201 & 0.481 & 0.705 & 0.872 \\
CLIP-L & 0.592 & 1.115 & -0.006 & -0.006 & 0.438 & 0.844 & 0.358 & 0.325 & 0.476 & 0.757 \\
BEiT-L & 0.583 & 1.190 & -0.060 & -0.053 & 0.408 & 0.908 & 0.217 & 0.465 & 0.682 & 0.858 \\
DINOv2-B & 0.574 & 1.107 & -0.054 & -0.052 & 0.403 & 0.862 & 0.305 & 0.378 & 0.554 & 0.796 \\
CLIP-B & 0.570 & 1.027 & 0.103 & 0.091 & 0.400 & 0.808 & 0.420 & 0.262 & 0.384 & 0.705 \\
EVA-02-B & 0.558 & 1.378 & -0.270 & -0.243 & 0.382 & 0.842 & 0.363 & 0.320 & 0.469 & 0.755 \\
BEiT-B & 0.546 & 1.461 & -0.331 & -0.293 & 0.360 & 0.872 & 0.298 & 0.384 & 0.563 & 0.801 \\
DeiT-3-L & 0.535 & 2.049 & -0.919 & -0.813 & 0.350 & 0.881 & 0.282 & 0.400 & 0.587 & 0.806 \\
SigLIP-2-L & 0.535 & 1.298 & -0.182 & -0.163 & 0.357 & 0.858 & 0.334 & 0.348 & 0.510 & 0.782 \\
AugReg-L & 0.532 & 1.607 & -0.477 & -0.422 & 0.341 & 0.837 & 0.358 & 0.324 & 0.475 & 0.731 \\
BEiT-3-B & 0.532 & 1.725 & -0.596 & -0.527 & 0.339 & 0.873 & 0.307 & 0.375 & 0.549 & 0.808 \\
DINOv3-B & 0.531 & 1.537 & -0.421 & -0.377 & 0.335 & 0.887 & 0.253 & 0.429 & 0.629 & 0.836 \\
AugReg-B & 0.530 & 1.546 & -0.417 & -0.369 & 0.336 & 0.843 & 0.346 & 0.336 & 0.492 & 0.743 \\
DINO-B & 0.527 & 1.466 & -0.337 & -0.298 & 0.340 & 0.870 & 0.312 & 0.370 & 0.542 & 0.798 \\
MAE-B & 0.526 & 1.904 & -0.774 & -0.685 & 0.341 & 0.839 & 0.362 & 0.320 & 0.469 & 0.768 \\
MAE-L & 0.525 & 1.970 & -0.840 & -0.743 & 0.335 & 0.866 & 0.306 & 0.376 & 0.551 & 0.802 \\
DeiT-3-B & 0.525 & 1.864 & -0.734 & -0.650 & 0.334 & 0.852 & 0.330 & 0.352 & 0.516 & 0.766 \\
SigLIP-2-B & 0.521 & 1.410 & -0.280 & -0.248 & 0.338 & 0.792 & 0.449 & 0.233 & 0.341 & 0.684 \\
FlexiViT-L & 0.514 & 1.820 & -0.697 & -0.621 & 0.309 & 0.854 & 0.363 & 0.320 & 0.468 & 0.762 \\
FlexiViT-B & 0.498 & 1.629 & -0.506 & -0.451 & 0.289 & 0.784 & 0.468 & 0.215 & 0.314 & 0.662 \\
\hline
Spatial prior & 0.496 & 1.093 & 0.000 & 0.000 & 0.314 & 0.561 & 0.683 & 0.000 & 0.000 & 0.370 \\
Area & 0.627 & -- & -- & -- & -- & 0.627 & -- & -- & -- & -- \\
Center overlap & 0.581 & -- & -- & -- & -- & 0.581 & -- & -- & -- & -- \\
Offset & 0.627 & -- & -- & -- & -- & 0.627 & -- & -- & -- & -- \\
\end{tabular}
\end{table}

\end{table}

\begin{figure}[!ht]
    \centering
    \includegraphics[width=0.9\linewidth]{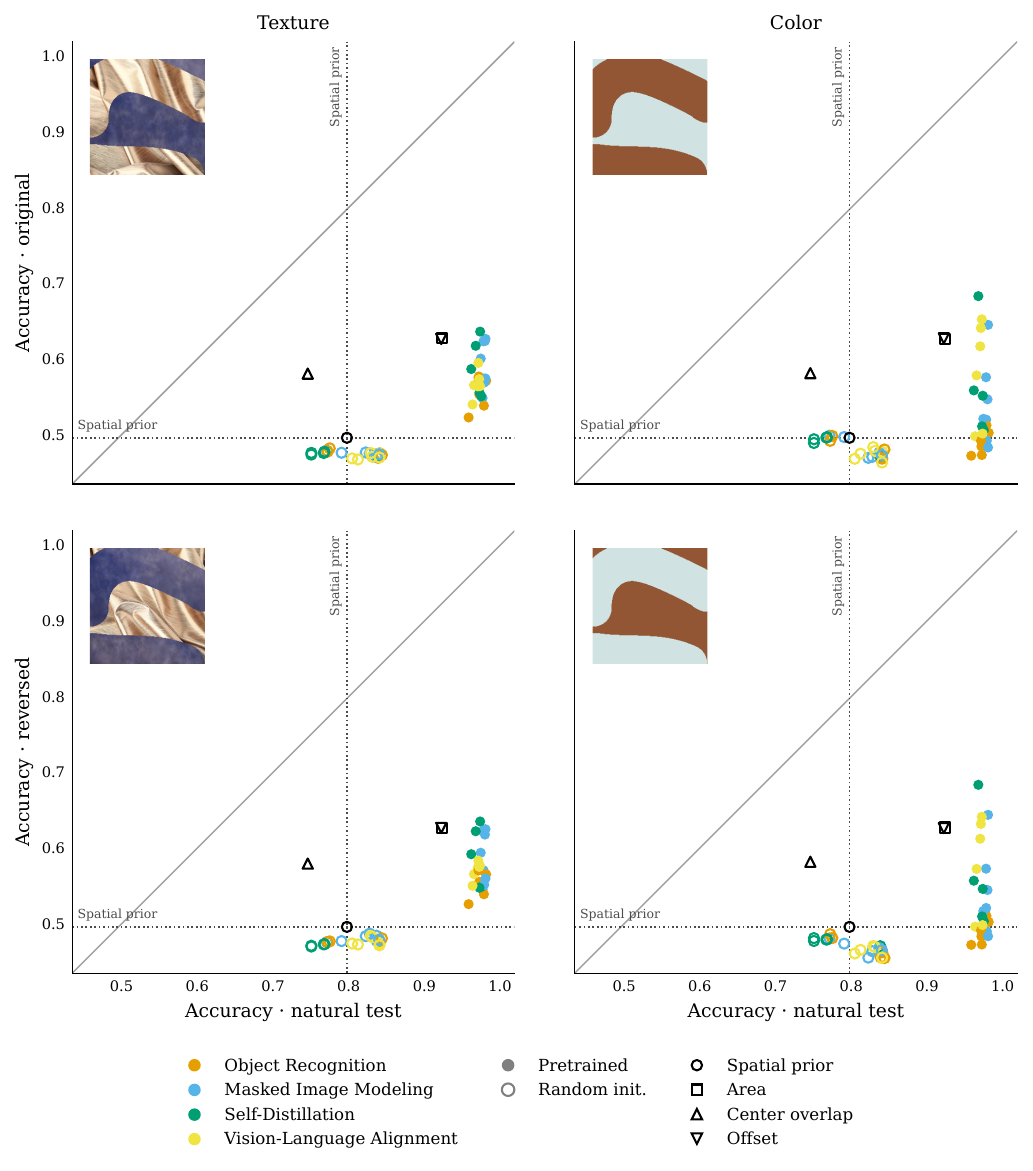}
    \caption{Comparison of the probe performances on the i.i.d. test set of natural images (x-axis) and applied zero-shot to the stimuli isolating parallelism (y-axis).}
    \label{fig:natural-vs-parallelism-variants}
\end{figure}

\clearpage
\begin{figure}[H]
    \centering
    \includegraphics[width=\linewidth]{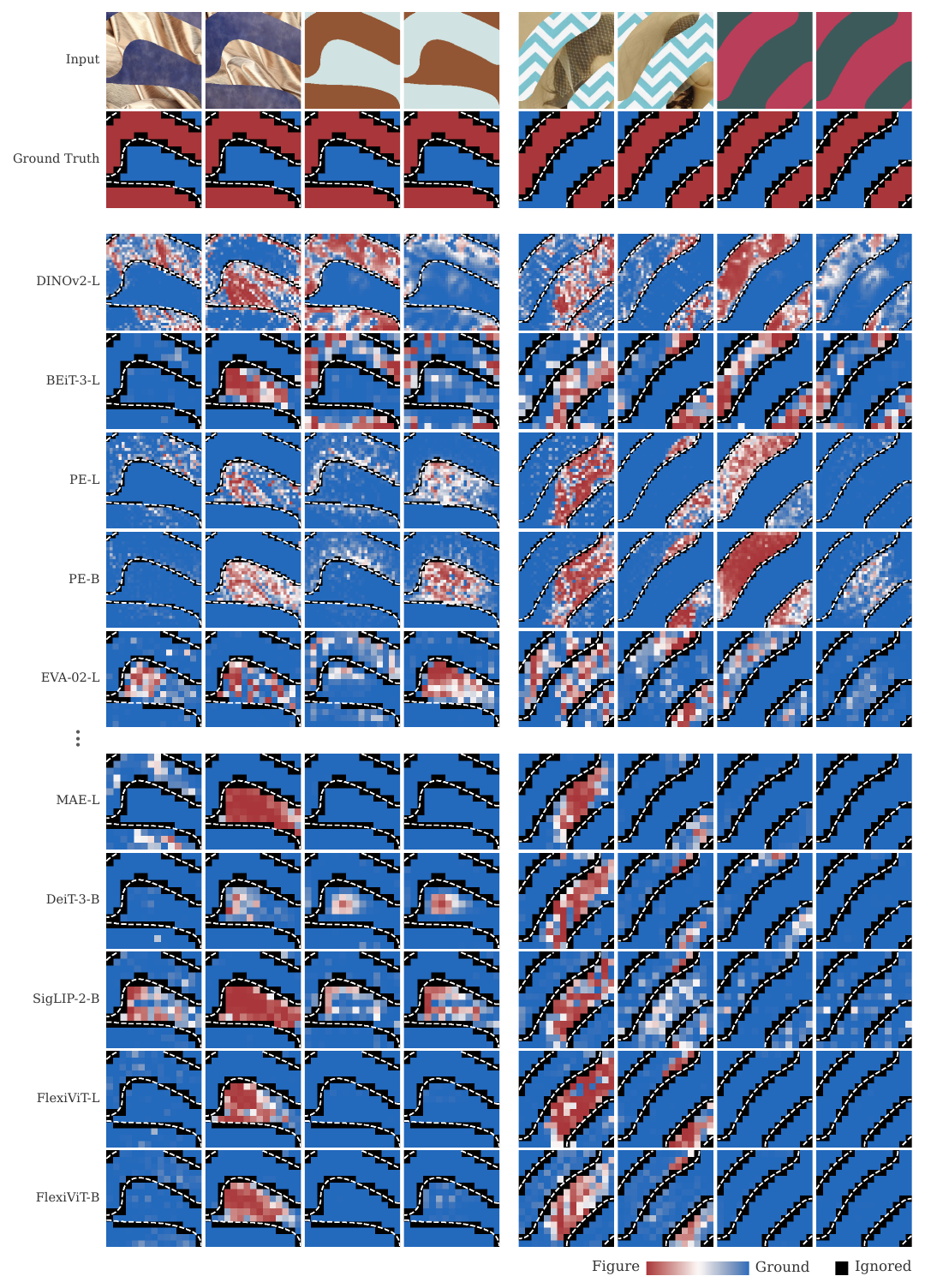}
    \caption{Probe predictions for the best and worst models in the parallelism condition.}
    \label{fig:predictions-parallelism}
\end{figure}

\clearpage
\begin{figure}[H]
    \centering
    \includegraphics[width=0.9\linewidth]{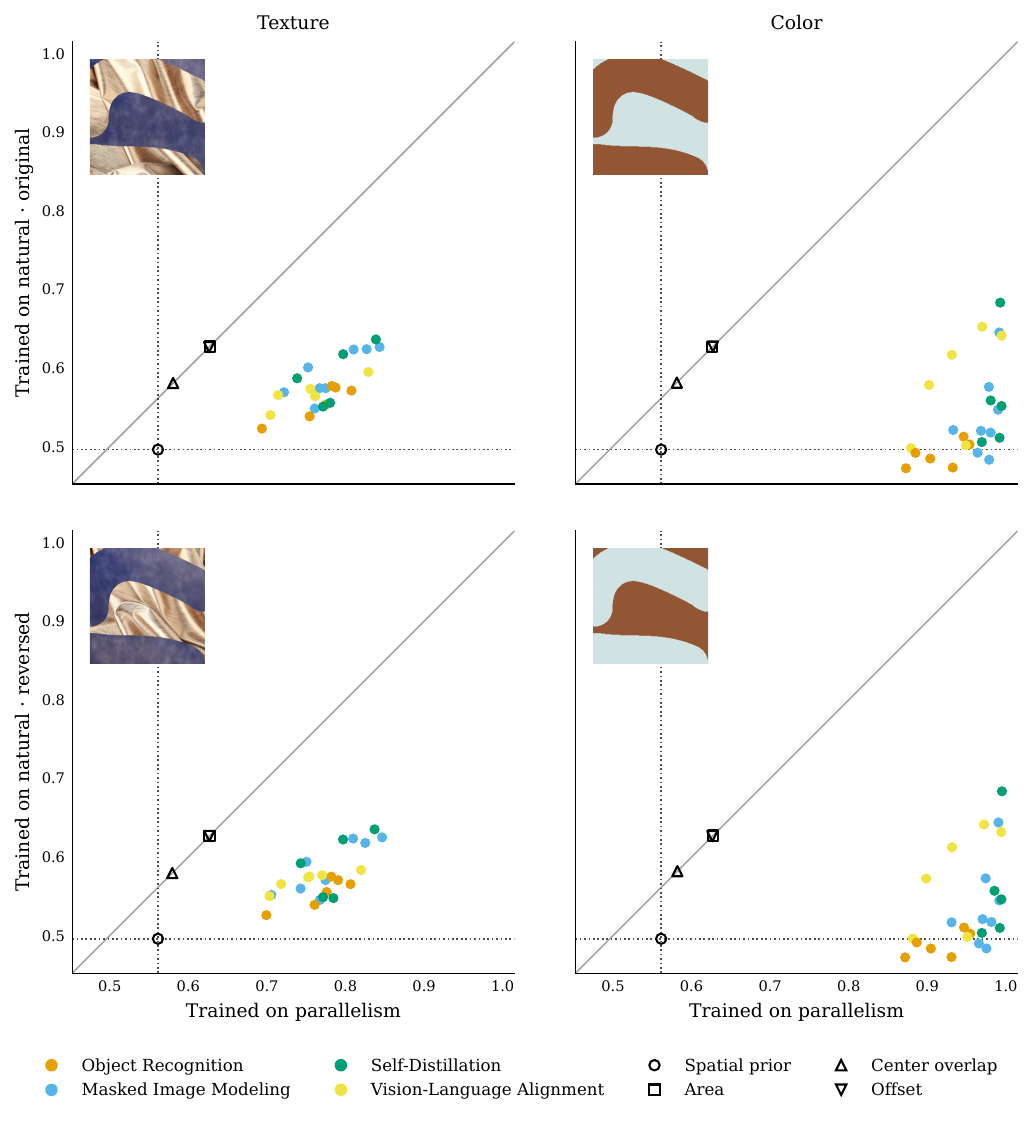}
    \caption{Performance comparison of probes directly trained on the parallelism stimuli (x-axis) with zero-shot probes trained on natural images (y-axis).}
    \label{fig:direct-training-parallelism-variants}
\end{figure}

\FloatBarrier

\section{Performance by Layer}
\begin{figure}[H]
    \centering
    \includegraphics[width=\linewidth]{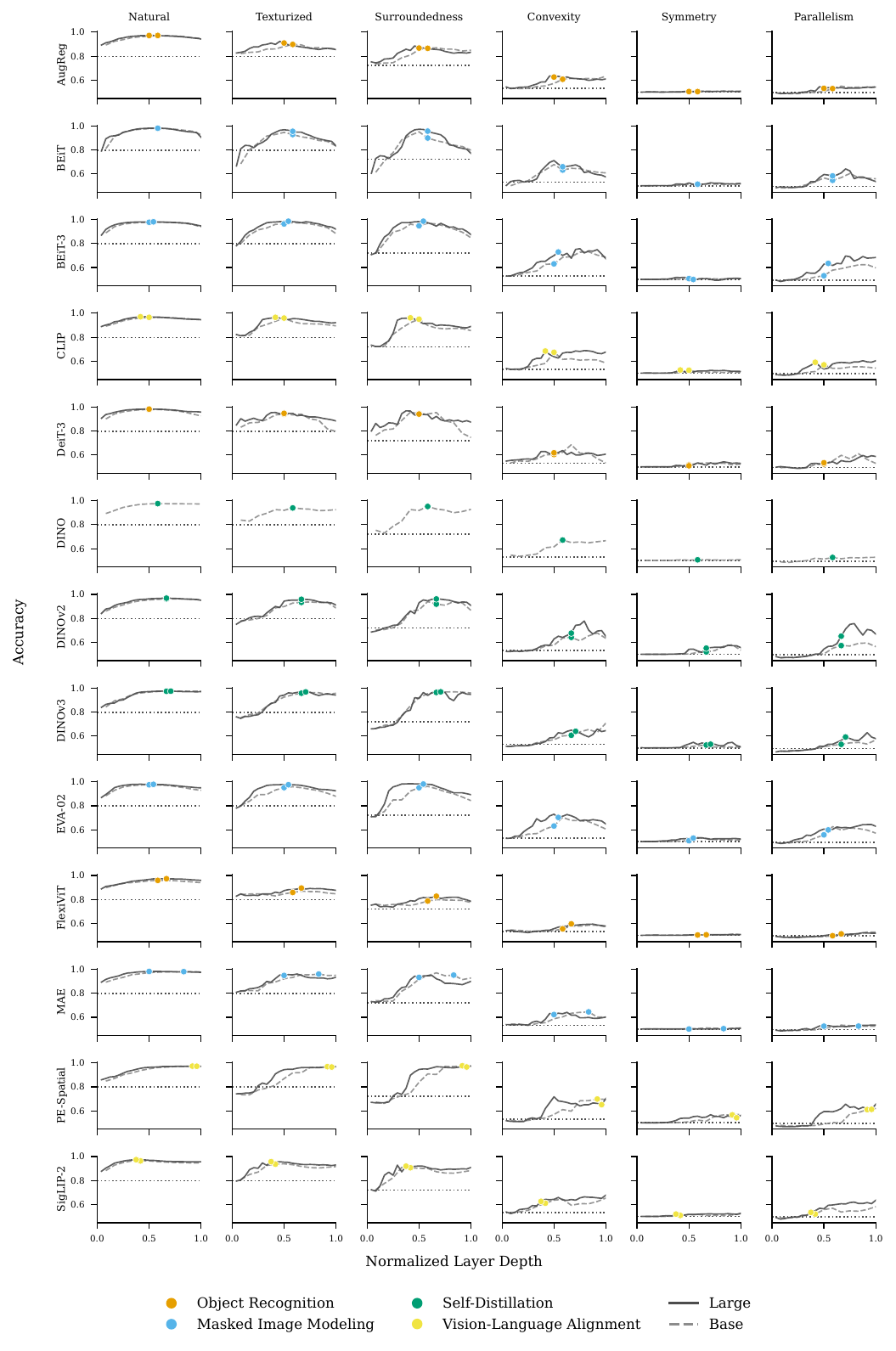}
    \caption{Individual probe performances for each model across layers.}
    \label{fig:performance-by-layer-details}
\end{figure}

\section{Detailed significance results for Figure \ref{fig:performance-by-size-and-task}}
\label{sec:appendix-significance-details}
\begin{table}[H]
\centering
\small
\caption{Summary statistics per condition. $W$: Wilcoxon signed-rank statistic (base vs. large, matched by model family); $r$: effect size $|Z|/\sqrt{n}$, with $Z$ the normal approximation of $W$ and $n$ the number of non-zero pairs; $H$: Kruskal-Wallis statistic (pretraining task); $\epsilon^2 = H/(N-1)$. Significance: $^{*}p<0.05$, $^{**}p<0.01$, $^{***}p<0.001$. \vspace{8pt}}
\label{tab:significance-overview}
\vspace{7pt}
\begin{tabular}{lrrllllrll}
Condition & N & N pairs & W & p(W) & r & H & df & p(H) & $\epsilon^2$ \\
\midrule
Natural & 25 & 12 & 4.0 & 0.003** & 0.793 & 11.744 & 3 & 0.008** & 0.489 \\
Texturized & 25 & 12 & 6.0 & 0.007** & 0.747 & 10.588 & 3 & 0.014* & 0.441 \\
Surroundedness & 25 & 12 & 13.0 & 0.042* & 0.589 & 9.116 & 3 & 0.028* & 0.380 \\
Convexity & 25 & 12 & 15.0 & 0.064 & 0.543 & 10.691 & 3 & 0.014* & 0.445 \\
Symmetry & 25 & 12 & 30.0 & 0.519 & 0.204 & 11.290 & 3 & 0.010* & 0.470 \\
Parallelism & 25 & 12 & 1.0 & <0.001*** & 0.861 & 6.380 & 3 & 0.095 & 0.266 \\
\end{tabular}
\end{table}

\begin{table}[H]
\centering
\small
\caption{Dunn post-hoc p-values (Holm-corrected) for all pretraining tasks. Significance: $^{*}p<0.05$, $^{**}p<0.01$, $^{***}p<0.001$. \vspace{8pt}}
\label{tab:significance-pretraining-task}
\vspace{7pt}
\begin{tabular}{lllllll}
Pair & Natural & Texturized & Surroundedness & Convexity & Symmetry & Parallelism \\
\midrule
MIM vs OR & 0.231 & 0.030* & 0.073 & 0.025* & 1.000 & 0.220 \\
OR vs SD & 1.000 & 0.242 & 0.062 & 0.094 & 0.123 & 0.206 \\
SD vs VLA & 1.000 & 1.000 & 1.000 & 1.000 & 1.000 & 1.000 \\
MIM vs SD & 0.082 & 1.000 & 1.000 & 1.000 & 0.123 & 1.000 \\
OR vs VLA & 0.672 & 0.020* & 0.073 & 0.027* & 0.058 & 0.168 \\
MIM vs VLA & 0.008** & 1.000 & 1.000 & 1.000 & 0.058 & 1.000 \\
\end{tabular}
\end{table}

\clearpage

\section{Licenses}
We use the DTD dataset \cite{cimpoi2014describing}, which does not provide a standard license but ``is made available to the computer vision community for research purposes.'' (\href{https://www.robots.ox.ac.uk/~vgg/data/dtd/}{https://www.robots.ox.ac.uk/~vgg/data/dtd/}, 2026-05-06).

Similarly, the MSRA-10K dataset \cite{cheng2015global} does not provide a standard license but requires researchers to cite their paper (\href{https://mmcheng.net/msra10k/}{https://mmcheng.net/msra10k/}, 2026-05-05).

We further use shapes from the Infinite DSprites dataset \cite{dziadzio2024infinite} which is licensed under the MIT license (\href{https://github.com/sbdzdz/idsprites/}{https://github.com/sbdzdz/idsprites/}, 2026-05-05).

The licenses and model cards for all models evaluated in our work are listed in Table~\ref{tab:model-info}.

\begin{table*}[!ht]
\centering
\caption{Vision Transformer checkpoints used in this work and their corresponding Hugging Face model cards and licenses.}
\small
\begin{tabular}{lll}
\textbf{Model} & \textbf{Checkpoint} & \textbf{License} \\
\hline
AugReg-B & \href{https://huggingface.co/timm/vit_base_patch16_224.augreg_in21k_ft_in1k}{timm/vit\_base\_patch16\_224.augreg\_in21k\_ft\_in1k} & Apache-2.0 \\
AugReg-L & \href{https://huggingface.co/timm/vit_large_patch16_224.augreg_in21k_ft_in1k}{timm/vit\_large\_patch16\_224.augreg\_in21k\_ft\_in1k} & Apache-2.0 \\
BEiT-3-B & \href{https://huggingface.co/timm/beit3_base_patch16_224.in22k_ft_in1k}{timm/beit3\_base\_patch16\_224.in22k\_ft\_in1k} & MIT \\
BEiT-3-BO & \href{https://huggingface.co/timm/beit3_base_patch16_224.pt}{timm/beit3\_base\_patch16\_224.pt} & MIT \\
BEiT-3-L & \href{https://huggingface.co/timm/beit3_large_patch16_224.in22k_ft_in1k}{timm/beit3\_large\_patch16\_224.in22k\_ft\_in1k} & MIT \\
BEiT-3-LO & \href{https://huggingface.co/timm/beit3_large_patch16_224.pt}{timm/beit3\_large\_patch16\_224.pt} & MIT \\
BEiT-B & \href{https://huggingface.co/timm/beit_base_patch16_224.in22k_ft_in22k}{timm/beit\_base\_patch16\_224.in22k\_ft\_in22k} & MIT \\
BEiT-L & \href{https://huggingface.co/timm/beit_large_patch16_224.in22k_ft_in22k}{timm/beit\_large\_patch16\_224.in22k\_ft\_in22k} & MIT \\
CLIP-B & \href{https://huggingface.co/timm/vit_base_patch16_clip_224.openai}{timm/vit\_base\_patch16\_clip\_224.openai} & MIT \\
CLIP-L & \href{https://huggingface.co/timm/vit_large_patch14_clip_224.openai}{timm/vit\_large\_patch14\_clip\_224.openai} & MIT \\
DINO-B & \href{https://huggingface.co/timm/vit_base_patch16_224.dino}{timm/vit\_base\_patch16\_224.dino} & Apache-2.0 \\
DINOv2-B & \href{https://huggingface.co/timm/vit_base_patch14_reg4_dinov2.lvd142m}{timm/vit\_base\_patch14\_reg4\_dinov2.lvd142m} & Apache-2.0 \\
DINOv2-L & \href{https://huggingface.co/timm/vit_large_patch14_reg4_dinov2.lvd142m}{timm/vit\_large\_patch14\_reg4\_dinov2.lvd142m} & Apache-2.0 \\
DINOv3-B & \href{https://huggingface.co/timm/vit_base_patch16_dinov3.lvd1689m}{timm/vit\_base\_patch16\_dinov3.lvd1689m} & DINOv3 License \\
DINOv3-L & \href{https://huggingface.co/timm/vit_large_patch16_dinov3.lvd1689m}{timm/vit\_large\_patch16\_dinov3.lvd1689m} & DINOv3 License \\
DeiT-3-B & \href{https://huggingface.co/timm/deit3_base_patch16_224.fb_in22k_ft_in1k}{timm/deit3\_base\_patch16\_224.fb\_in22k\_ft\_in1k} & Apache-2.0 \\
DeiT-3-L & \href{https://huggingface.co/timm/deit3_large_patch16_224.fb_in22k_ft_in1k}{timm/deit3\_large\_patch16\_224.fb\_in22k\_ft\_in1k} & Apache-2.0 \\
EVA-02-B & \href{https://huggingface.co/timm/eva02_base_patch14_224.mim_in22k}{timm/eva02\_base\_patch14\_224.mim\_in22k} & MIT \\
EVA-02-L & \href{https://huggingface.co/timm/eva02_large_patch14_224.mim_in22k}{timm/eva02\_large\_patch14\_224.mim\_in22k} & MIT \\
FlexiViT-B & \href{https://huggingface.co/timm/flexivit_base.1200ep_in1k}{timm/flexivit\_base.1200ep\_in1k} & Apache-2.0 \\
FlexiViT-L & \href{https://huggingface.co/timm/flexivit_large.1200ep_in1k}{timm/flexivit\_large.1200ep\_in1k} & Apache-2.0 \\
MAE-B & \href{https://huggingface.co/timm/vit_base_patch16_224.mae}{timm/vit\_base\_patch16\_224.mae} & CC-BY-NC-4.0 \\
MAE-L & \href{https://huggingface.co/timm/vit_large_patch16_224.mae}{timm/vit\_large\_patch16\_224.mae} & CC-BY-NC-4.0 \\
PE-B & \href{https://huggingface.co/timm/vit_pe_spatial_base_patch16_512.fb}{timm/vit\_pe\_spatial\_base\_patch16\_512.fb} & Apache-2.0 \\
PE-L & \href{https://huggingface.co/timm/vit_pe_spatial_large_patch14_448.fb}{timm/vit\_pe\_spatial\_large\_patch14\_448.fb} & Apache-2.0 \\
SigLIP-2-B & \href{https://huggingface.co/timm/vit_base_patch16_siglip_224.v2_webli}{timm/vit\_base\_patch16\_siglip\_224.v2\_webli} & Apache-2.0 \\
SigLIP-2-L & \href{https://huggingface.co/timm/vit_large_patch16_siglip_256.v2_webli}{timm/vit\_large\_patch16\_siglip\_256.v2\_webli} & Apache-2.0 \\
\end{tabular}
\label{tab:model-info}
\end{table*}

\clearpage
\newpage
\section*{NeurIPS Paper Checklist}

\begin{enumerate}

\item {\bf Claims}
    \item[] Question: Do the main claims made in the abstract and introduction accurately reflect the paper's contributions and scope?
    \item[] Answer: \answerYes{}
    \item[] Justification: We provide experimental results for all claims in the abstract and introduction.

\item {\bf Limitations}
    \item[] Question: Does the paper discuss the limitations of the work performed by the authors?
    \item[] Answer: \answerYes{}
    \item[] Justification: The paper includes a dedicated section on the limitations of our work.

\item {\bf Theory assumptions and proofs}
    \item[] Question: For each theoretical result, does the paper provide the full set of assumptions and a complete (and correct) proof?
    \item[] Answer: \answerNA{}
    \item[] Justification: The paper does not contribute theoretical results.

    \item {\bf Experimental result reproducibility}
    \item[] Question: Does the paper fully disclose all the information needed to reproduce the main experimental results of the paper to the extent that it affects the main claims and/or conclusions of the paper (regardless of whether the code and data are provided or not)?
    \item[] Answer: \answerYes{}
    \item[] Justification: We carefully describe all details needed to reproduce our results in the method section, including details on stimulus generation, probe training and evaluation. Additionally, we publish all code for stimulus generation, probe training and evaluation.

\item {\bf Open access to data and code}
    \item[] Question: Does the paper provide open access to the data and code, with sufficient instructions to faithfully reproduce the main experimental results, as described in supplemental material?
    \item[] Answer: \answerYes{}
    \item[] Justification: All code and data for this is publicly available under the link provided in the abstract.

\item {\bf Experimental setting/details}
    \item[] Question: Does the paper specify all the training and test details (e.g., data splits, hyperparameters, how they were chosen, type of optimizer) necessary to understand the results?
    \item[] Answer: \answerYes{}{}
    \item[] Justification: The methods section describes our experimental setup in great detail.

\item {\bf Experiment statistical significance}
    \item[] Question: Does the paper report error bars suitably and correctly defined or other appropriate information about the statistical significance of the experiments?
    \item[] Answer: \answerYes{}
    \item[] Justification: We evaluate and report statistical significance for our claims regarding the influence of model size and pre-training task. Evaluating the statistical significance of the model ranking for each condition is prohibitive given the compute requirements. We therefore do not make any strong claims about detailed model rankings.

\item {\bf Experiments compute resources}
    \item[] Question: For each experiment, does the paper provide sufficient information on the computer resources (type of compute workers, memory, time of execution) needed to reproduce the experiments?
    \item[] Answer: \answerYes{}
    \item[] Justification: We report the GPU type used (NVIDIA L40S) and the total time required for fitting the probes (120 GPU hours).
    
\item {\bf Code of ethics}
    \item[] Question: Does the research conducted in the paper conform, in every respect, with the NeurIPS Code of Ethics \url{https://neurips.cc/public/EthicsGuidelines}?
    \item[] Answer: \answerYes{}
    \item[] Justification: We comply with the NeurIPS Code of Ethics.

\item {\bf Broader impacts}
    \item[] Question: Does the paper discuss both potential positive societal impacts and negative societal impacts of the work performed?
    \item[] Answer: \answerNA{}
    \item[] Justification: We study the internal processing of published models from the perspective of research of human vision. We don't see any potential negative societal impacts of this work.
    
\item {\bf Safeguards}
    \item[] Question: Does the paper describe safeguards that have been put in place for responsible release of data or models that have a high risk for misuse (e.g., pre-trained language models, image generators, or scraped datasets)?
    \item[] Answer: \answerNA{}
    \item[] Justification: We don't see any such risks.

\item {\bf Licenses for existing assets}
    \item[] Question: Are the creators or original owners of assets (e.g., code, data, models), used in the paper, properly credited and are the license and terms of use explicitly mentioned and properly respected?
    \item[] Answer: \answerYes{}
    \item[] Justification: We cite all papers for datasets and models as required by the respective authors. Furthermore, we list licenses and model cards in the appendix.

\item {\bf New assets}
    \item[] Question: Are new assets introduced in the paper well documented and is the documentation provided alongside the assets?
    \item[] Answer: \answerNA{}
    \item[] Justification: We do not provide a new model and did not collect a new dataset.

\item {\bf Crowdsourcing and research with human subjects}
    \item[] Question: For crowdsourcing experiments and research with human subjects, does the paper include the full text of instructions given to participants and screenshots, if applicable, as well as details about compensation (if any)? 
    \item[] Answer: \answerNA{}
    \item[] Justification: Our work does not include experiments with human subjects.

\item {\bf Institutional review board (IRB) approvals or equivalent for research with human subjects}
    \item[] Question: Does the paper describe potential risks incurred by study participants, whether such risks were disclosed to the subjects, and whether Institutional Review Board (IRB) approvals (or an equivalent approval/review based on the requirements of your country or institution) were obtained?
    \item[] Answer: \answerNA{} 
    \item[] Justification: Our work does not include experiments with human subjects.

\item {\bf Declaration of LLM usage}
    \item[] Question: Does the paper describe the usage of LLMs if it is an important, original, or non-standard component of the core methods in this research? Note that if the LLM is used only for writing, editing, or formatting purposes and does \emph{not} impact the core methodology, scientific rigor, or originality of the research, declaration is not required.
    \item[] Answer: \answerNo{}
    \item[] Justification: The core method of our work does not involve LLM usage.

\end{enumerate}

\end{document}